\pdfoutput=1

\documentclass[11pt]{article}

\usepackage{acl}

\usepackage{times}
\usepackage{latexsym}

\usepackage{wrapfig}

\usepackage[T1]{fontenc}

\usepackage[utf8]{inputenc}

\usepackage{microtype}

\usepackage{inconsolata}

\usepackage{graphicx}
\usepackage{adjustbox}
\usepackage{amsmath}

\title{What Do VLMs NOTICE? A Mechanistic Interpretability Pipeline for Gaussian-Noise-free Text-Image Corruption and Evaluation}

\author{Michal Golovanesky $^{\star 1}$, William Rudman $^{\star 1}$, \
Vedant Palit$^{2}$, Ritambhara Singh$^{1,3}$, Carsten Eickhoff$^{4}$ \\ 
  $^{1}$Department of Computer Science, Brown University \\
  $^{2}$Indian Institute of Technology Kharagpur\\
  $^{3}$Center for Computational Molecular Biology, Brown University\\
  $^{4}$School of Medicine, University of Tübingen \\
\texttt{\{michal\_golovanevsky, william\_rudman\}@brown.edu} \\
  }

\date{}

\begin{document}

\maketitle

\renewcommand{\thefootnote}{\fnsymbol{footnote}}
\footnotetext[1]{Authorship order determined by coin flip.}

\begin{abstract}
Vision-Language Models (VLMs) have gained prominence due to their success in solving complex cross-modal tasks. However, the internal mechanisms of VLMs, particularly the roles of cross-attention and self-attention in multimodal integration, are not fully understood. To address this gap, we introduce NOTICE, a Gaussian-\textbf{NO}ise-free \textbf{T}ext-\textbf{I}mage \textbf{C}orruption and \textbf{E}valuation pipeline for mechanistic interpretability in VLMs. NOTICE introduces Semantic Image Pairs (SIP) corruption, the first visual counterpart to Symmetric Token Replacement (STR) for text. Through NOTICE, we uncover a set of ``universal attention heads'' in BLIP and LLaVA that consistently contribute across different tasks and modalities. In BLIP, cross-attention heads implement object detection, object suppression, and outlier suppression, whereas important self-attention heads in LLaVA only perform outlier suppression. Notably, our findings reveal that cross-attention heads perform image-grounding, while self-attention in LLaVA heads do not, highlighting key differences in how VLM architectures handle multimodal learning\footnote{Code: 
\href{https://github.com/wrudman/NOTICE/}{
\textit{https://github.com/wrudman/NOTICE/}}}. 


\end{abstract}

\section{Introduction}


Vision-Language Models (VLMs) have become central to both computer vision and natural language processing (NLP) due to their ability to complete complex multimodal tasks like visual recognition \citep{han2023llms, menon2022visual}, visual question answering \citep{liu2024visual, bai2023qwen, li2022blip, li2020oscar}, and image captioning \citep{lu2019vilbert, yang2023vid2seq, zhang2021vinvl}. Many foundational VLMs, such as BLIP (Bootstrapping Language-Image Pre-training) \citep{li2022blip}, use cross-attention to capture interactions between modalities \cite{tan2019lxmert, lu2019vilbert, li2022blip}, while others, like LLaVA \cite{liu2024visual}, successfully use self-attention for multimodal integration. Despite recent works investigating the differences in cross-attention and self-attention in VLMs \citep{attention_as_grounding, parcalabescu2020seeing, djamila2022automatic}, the specific role of different attention heads in vision-language is not fully understood, particularly in large VLMs such as LLaVA \cite{yuksekgonul2022and}.

\begin{figure*}[h!]
    \begin{center}
        \includegraphics[width=\textwidth]{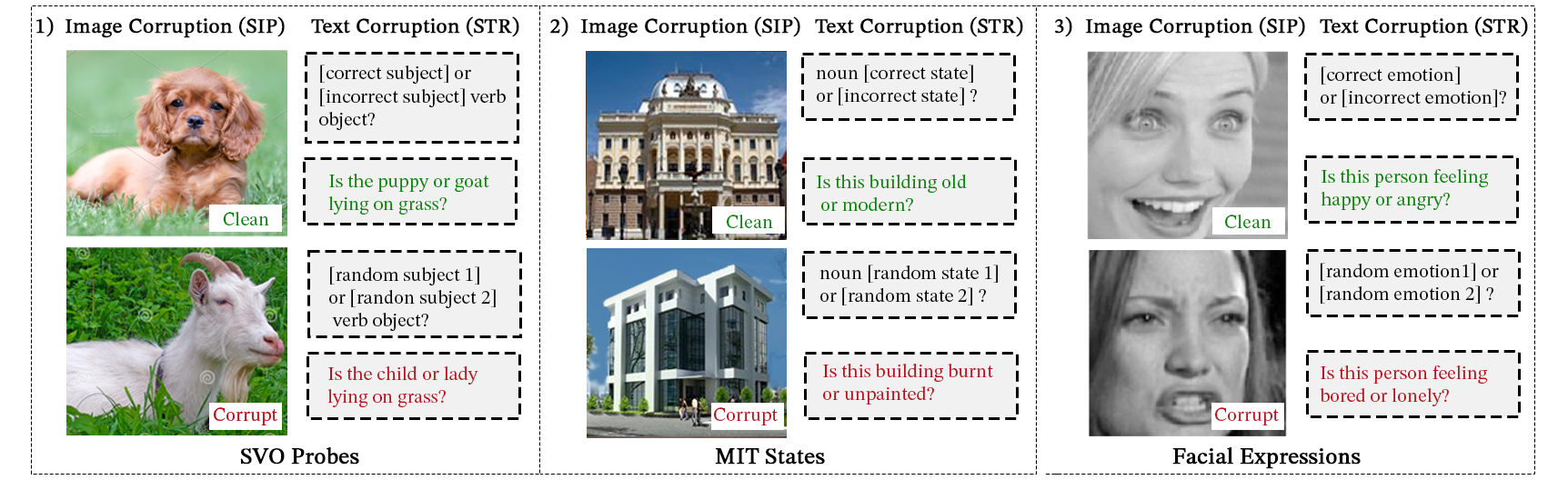} 
        \caption{NOTICE applied to SVO-Probes, MIT-States, and Facial Expression Recognition. NOTICE involves creating Semantic Image Pairs for image corruption and Symmetric Token Replacement for text corruption.}
        \label{fig:corruption_scheme}
    \end{center}
\end{figure*}

In NLP, mechanistic interpretability (MI) has proven effective for unraveling the complex structure of LLMs by identifying pathways, or \textit{circuits}, that enable behaviors like logical reasoning and indirect object identification \cite{gpt2_greater_than, wang2022interpretability}. A core method in MI is \textit{causal mediation analysis} (CMA),  which measures output changes by intervening on inputs and tracking effects. To apply CMA in Transformers, \citet{meng2023locating} use \textit{activation patching}, which identifies key model components for a given task by replacing the hidden states of a \textit{corrupt} run with those from a \textit{clean} run. A ``clean run'' refers to a standard forward pass on an input. A ``corrupt'' run either alters the input by swapping input tokens with semantically related tokens or applies Gaussian noise to token embeddings so that the model can no longer produce the correct answer from the clean pass. While successful in NLP, mechanistic interpretability methods are understudied in VLMs due to the need for an effective semantic-based corruption scheme for images. 

Previous works applying activation patching to VLMs only study BLIP, focus on Multi-Layer Perceptron (MLP) layers in BLIP's decoder, and apply Gaussian noise to image embeddings to create corrupt runs \cite{Palit_2023_ICCV}. However, multimodal interaction occurs in BLIP's image-grounded text encoder through the cross-attention layers, meaning findings from \citet{Palit_2023_ICCV} do not provide insights on how VLMs integrate modalities. Further, recent works have shown that creating corrupt runs by applying Gaussian noise to token embeddings can produce illusory results \cite{best_practices}. \citet{best_practices} advocate for \textit{symmetric token replacement} (STR) that swaps tokens in a clean prompt with semantically related tokens to create a corrupt run as it keeps the corrupted run within distribution and provides more reliable insights Current approaches to image corruption lack an equivalent method to STR that does not use Gaussian noise for image corruption, leaving a gap in reliable multimodal analysis. 

We propose \textit{Semantic Image Pairs} (SIP) corruption, which leverages existing datasets to create clean and corrupt images that vary by one semantic property (Figure \ref{fig:corruption_scheme}). For instance, in the SVO-Probes dataset \citep{hendricks2021probing}, subject-verb-object triplets are already curated to vary by just one element (e.g., 'puppy, lying, grass' vs. 'goat, lying, grass'), allowing us to create ``clean'' and ``corrupt'' image pairs from the existing image labels (Figure \ref{fig:corruption_scheme} panel 1). SIP is not limited to curated datasets and can be extended to any image dataset with easily interchangeable attributes, such as cars, shapes, animals, etc. Further, we show in Section~\ref{sec:module-wise} that SIP corruption can be performed using generative models to create corrupt images. 

Our framework, NOTICE (Gaussian-\textbf{NO}ise-free \textbf{T}ext-\textbf{I}mage \textbf{C}orruption and \textbf{E}valuation), is a mechanistic interpretability pipeline for VLMs. NOTICE uses our novel SIP method to corrupt images and STR to corrupt text. Using NOTICE, we uncover ``universal cross-attention heads'' in BLIP and ``universal self-attention heads'' in LLaVA that consistently produce large patching effects across all tasks. We show that universal cross-attention heads implement human-interpretable functions categorized into three classes: object detection, object suppression, and outlier suppression. We find that both self-attention and cross-attention heads implement outlier suppression. However, only cross-attention heads in BLIP perform the image-grounding functions of object detection and object suppression, highlighting key differences in how VLM architectures handle multimodal learning. 

\section{Related Work}

\textbf{Multimodal Interpretability}. Despite the successes of VLMs, the role of attention in multimodal integration remains under-studied \cite{multimodal_learning_survey}. Most works either design probing tasks to identify what vision-language concepts models struggle to capture \citep{foil, hendricks2021probing, vim_probing} or assess whether models rely more on one modality over the other \cite{behind_the_scenes, are_vl_transformers_multimodal, emap}. Probing tasks evaluate whether a model encodes a concept by training a linear classifier on its hidden states. \citet{are_vl_transformers_multimodal} create visual, text, and multimodal probing tasks to show that models heavily rely on text priors and struggle with object, size, and counting. Gradient-based methods analyze gradients to understand how changes in input features influence outputs, assigning importance scores to features and providing insights into the decision-making process. \citet{multiviz} use second-order gradients, building on LIME \cite{ribeiro2016should} and Shapley values \cite{merrick2020explanation}, to disentangle cross-modal interactions and determine how much each modality contributes to decisions. While these works have provided insights into multimodal models, recent research shows that 1) probing methods can produce misleading results \citep{probing_limitations} and 2) gradient-based methods for feature attributions can fail to predict behavior \citep{impossibility_theorems}. Instead, \citet{giulianelli2021hood,conneau-etal-2018-cram} demonstrate that causal approaches are more informative than probing representations.

\begin{figure*}[h]
\begin{centering}
    \includegraphics[width=\textwidth]{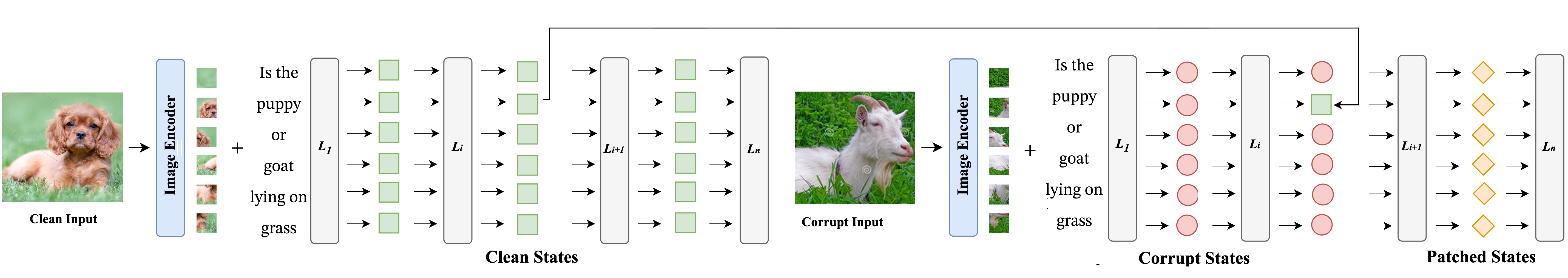}
    \caption{Activation Patching using SIP corruption. The image of the puppy is the clean image, $I$, and the goat is the corrupt image, $I^{*}$. Patching the correct answer token ``puppy'' at $M_{l}(I,T)$, from the clean image to the ``puppy'' token at $M(I^{*}, T)$ creates the patched states $M^{'}(I^{*}, T)$ shown as orange diamonds.}
    \label{fig:patching_diagram}
\end{centering}

\end{figure*}

Multimodal models typically integrate their inputs through one of two main approaches: (1) early fusion \cite{liu2024visual, li2019visualbert}, where self-attention processes combined image patches and text representations, or (2) cross-attention fusion \cite{tan2019lxmert, lu2019vilbert}, where queries from one modality interact with representations from the other. 
\cite{frank2021vision} investigate cross-modal integration using input ablation techniques and conclude that self-attention models, such as VisualBERT, show less sensitivity to missing visual input compared to cross-attention models like ViLBERT, which exhibit more significant reliance on cross-modal interactions.\cite{ilinykh2022attention} finds that vision-text cross-attention learns visual grounding of noun phrases into objects, while text-to-text attention captures low-level syntactic word relations.\cite{parcalabescu2020seeing} probes dual-stream VLMs through tasks like image-sentence alignment and counting, revealing strong alignment performance but limited grounding of visual symbols, resulting in poor counting capabilities. However, there has yet to be a comprehensive study applying mechanistic interpretability techniques to investigate how recent architectures like BLIP and LLaVA 1) use attention mechanisms to integrate multimodal information and 2) determine the specific functions that attention heads in these models learn to perform. 

\textbf{Causal Mediation Analysis and Mechanistic Interpretability.} Causal mediation analysis (CMA) uncovers how interventions in one part of a system cause changes in another. Recent works have adapted CMA to study language models by using \textit{activation patching} \cite{meng2023locating} to identify causal pathways, or \textit{circuits}, responsible for certain behaviors in the model. Activation patching shows the indirect effect of a component on a model's output by corrupting key input tokens and then restoring outputs by patching the internal modules from the clean run to the corrupted run. This approach has revealed that attention heads in transformers implement human-interpretable algorithms for ``greater than'' tasks \cite{gpt2_greater_than}, indirect object identification \cite{wang2022interpretability}, and learning mathematical group operations \citep{circuits_groups}. Evidence suggests that models implement similar circuits across various tasks, indicating that studying a small set of computational units may provide insights into model behavior \citep{successor_heads, circuit_reuse}. Although works in mechanistic interpretability have provided critical insights into LLMs, works applying activation patching to multimodal models have been limited due to the absence of an effective semantic-based method for image corruption. Given that \citet{best_practices} demonstrate Gaussian noise corruption can lead to ``illusory'' patching results, a semantic-based corruption scheme for images is needed to apply causal mediation analysis to VLMs. 
In this work, we introduce a semantic-based image corruption method (SIP) and token replacement (STR) for text, enabling meaningful causal mediation analysis to reveal distinct roles of self-attention and cross-attention in multimodal models like BLIP and LLAVA, advancing our understanding of vision-language integration.

\section{Methods}
\textbf{NOTICE: Overview} The first step in activation patching is developing a method to corrupt the input, preventing the model from predicting the correct answer without causing representations to be out-of-distribution. We propose the first semantic-based dual corruption scheme for VLMs to enhance understanding of each modality's contribution to VQA tasks. We assess the impact of corrupting input text with symmetric token replacement \cite{wang2022interpretability} and images using our novel \textit{Semantic Image Pairs} (SIP) framework.

\textbf{Model Selection} For our model selection, we use LLaVA-1.5-7B (Vicuna base) \cite{liu2024visual} and BLIP-VQA-Base \cite{li2022blip}, chosen for their relevance and alignment with recent interpretability research. BLIP-VQA was included to enable direct comparison with prior work applying activation patching to vision-language models \cite{Palit_2023_ICCV}, facilitating a meaningful evaluation of our corruption scheme (NOTICE) against Gaussian-noise patching. In addition, we also focus on LLaVA-1.5 to align with more recent interpretability studies, including \cite{luo2024task}, \cite{neo2024towards}, and \cite{jiang2024interpreting}.

    
\textbf{Text Corruption: Symmetric Token Replacement.} Corrupting text inputs with symmetric token replacement (STR) swaps out the correct answer token of a prompt with another similar token that has the same length \cite{wang2022interpretability}. For indirect object identification, \citet{wang2022interpretability} swap the names of the indirect and direct objects with randomly sampled names. The goal of STR is to corrupt inputs so the language model cannot complete the task correctly while avoiding Gaussian noise corruption that pushes token embeddings off-distribution. For our VQA tasks, we swap both multiple-choice options in the input prompt with two randomly sampled incorrect multiple-choice options with the same tokenized length from a different sample in the dataset. Figure~\ref{fig:corruption_scheme} illustrates how STR alters the input text for a prompt.

\begin{figure*}[h!]
    \centering
    \includegraphics[width=\textwidth]{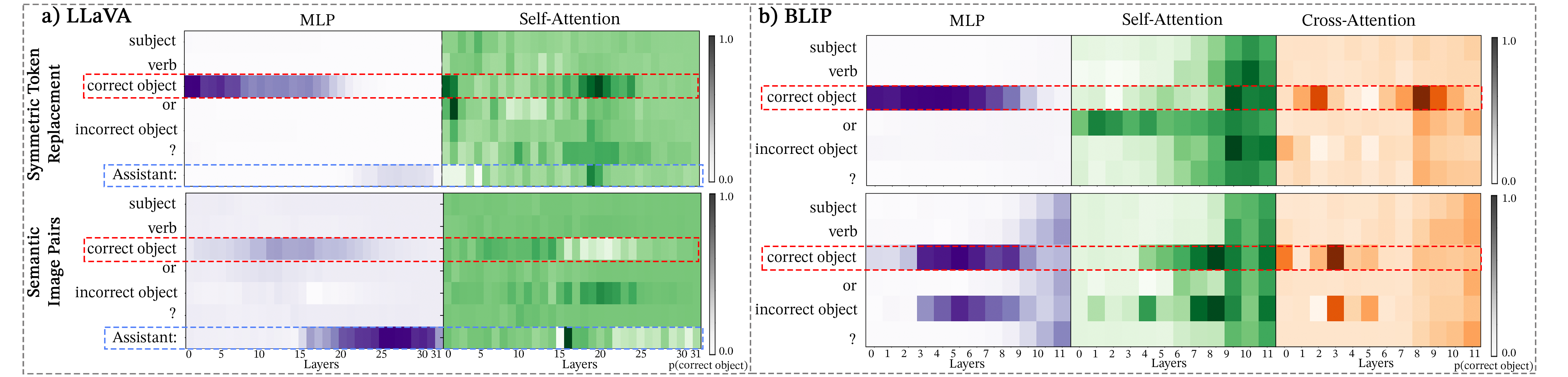}
    \caption{Module-wise activation patching results for BLIP and LLaVA on ``objects'' from SVO-Probes. We visualize the restoration probability after patching for MLP, self-attention, and cross-attention layers in the image-grounded text-encoder. The y-axis denotes which token we patch, and the x-axis denotes which layer we patch.}
    \label{fig:blip_llava}
\end{figure*}

\textbf{Image Corruption: Semantic Image Pairs.} While STR is a widely used text corruption method, extending semantic corruption to the vision domain presents significant challenges as it requires a semantically aligned image pair. To address this, we propose SIP corruption, a visual corollary to STR, which modifies only one semantic concept in an image (see Figure~\ref{fig:corruption_scheme}).
We implement SIP in three diverse image-classification datasets: SVO-Probes \citep{hendricks2021probing} (an extension to FOIL it! \cite{shekhar2017foil}), MIT States \citep{isola2015discovering}, and Facial Expressions \citep{goodfellow2013challenges}. While natural variation in backgrounds can happen between semantic image pairs, we select datasets where nouns, subjects, verbs, and objects are known and can be controlled. The Facial Expressions dataset was curated to include only black-and-white photos of faces expressing different emotions. In MIT-States, object-state pairs are known and can be used to maintain objects (e.g., ``building'') while varying their states (e.g., ``modern''). The SVO-Probes dataset is designed to vary only one element (subject, verb, or object) at a time (e.g., only varying ``puppy'' and ``goat'', while maintaining ``grass'' and ``lying'' constant, as seen in Figure \ref{fig:corruption_scheme} panel 1). Thus, for each dataset, we collect the known elements (emotions, objects/states, subjects/verbs/objects) and create image pairs that differ on that element. Our framework can be extended to any image dataset with clearly defined attributes, such as cars, shapes, animals, etc., using the class labels to find clean and corrupt pairs. 

Previous works have shown vision transformers can correctly identify objects in various contexts regardless of the foreground or background  \cite{vilas2024analyzing}. Since image encoders in multimodal models, such as BLIP and LLaVA, are robust to minor variations to the context of an object, only the semantic contents of the image need to be controlled to obtain reliable patching results \cite{vilas2024analyzing}. In contrast, masking pixels, adding Gaussian noise to image patches, or replacing pixels would create unnatural images, leading to unreliable results. In Section \ref{sec:module-wise}, we demonstrate the effectiveness of SIP corruption over Gaussian-noise image corruption and experiment with using generative models to create SIP. 

\textbf{NOTICE VQA Task Design.} We adapt the SVO-Probes, MIT States, and Facial Expressions image-classification datasets into VQA tasks. We prompt models to choose between a correct answer that matches the image and an incorrect, semantically different alternative, as seen in Figure \ref{fig:corruption_scheme}. For example, using the image-class labels ``happy'' and ``angry,'' we create a VQA prompt: ``Is this person feeling happy or angry?''. For LLaVA, we adapt the standard prompting format to ensure the model selects one of the multiple choice options: ``USER: <image> [Prompt]. Answer with one word. Assistant:''. While in SVO-Probes, each subject, verb, or object comes with its own semantically different counterpart, we constructed such pairs for MIT-States and Facial Expression. In MIT-States, we separated states into color, shape, and texture and selected counterparts from the respective categories for each question. We grouped similar states (e.g., ``huge'', ``large'') together and ensured that the selected counterpart would not be similar in meaning to the correct answer (using cosine similarity and word2vec word embeddings \cite{word2vec}). Similarly, in the Facial Expressions dataset, we grouped the seven emotions into positive, negative, and neutral and selected counterparts from opposing emotions. We include more details on the VQA design and the datasets in Appendix~\ref{app:vqa_prompts} and Appendix~\ref{app:datasets}, respectively. 

\textbf{Activation Patching.} Let $M$ be a vision-language model and $M_{l}$ its hidden states at layer $l \in \{0,1,...,n\}$. The forward pass on an image ($I$) and text ($T$) pair at layer $l$ is denoted as $M_{l}(I,T)$. We define $I^{*}$ as a corrupted image and $T^{*}$ as a corrupted text input. Note that we only corrupt one modality at a time. In a forward pass with $I^{*}$ and $T$, activation patching replaces the hidden states of the corrupted run $M_{l}(I^{*}, T)$ with the clean run $M_{l}(I,T)$ to determine which components have the most significant indirect effect on the output. The hidden states at all subsequent layers, $M^{'}_{j}(I^{*}, T)$ for $j \in \{l+1,...,n\}$, are called \textit{patched states}. In Figure~\ref{fig:patching_diagram}, we provide an example illustrating patching using SIP image corruption. The image of the puppy is the clean image, $I$, and the goat is $I^{*}$. Patching the correct answer token ``puppy'' at $M_{l}(I,T)$, from the clean image to the ``puppy'' token at $M(I^{*}, T)$ creates the patched states $M^{'}(I^{*}, T)$ shown as orange diamonds in Figure~\ref{fig:patching_diagram}.

\begin{figure*}[h!]
    \centering
    \includegraphics[width=16cm]{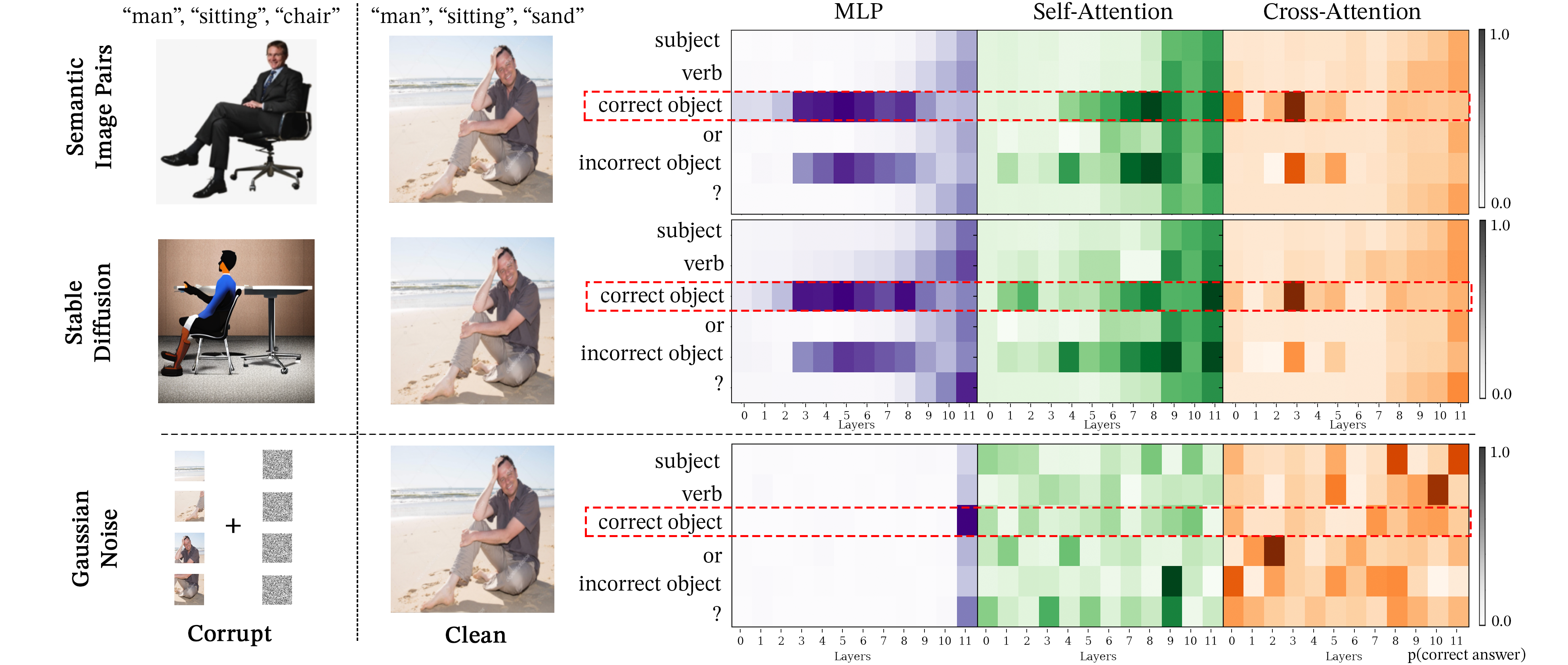}
    \caption{Module-wise activation patching results for SIP, and Gaussian-Noise corruption on SVO-Probes on BLIP. SIP corruption produces activation patterns that align with Stable Diffusion results and highlight the importance of middle layers, while Gaussian noise fails to reveal meaningful attention layers, emphasizing the effectiveness of SIP for probing vision-language models.}
    \label{fig:blip_svo_noise}
\end{figure*}

\textbf{Measuring Patching Effects.} The most common ways of measuring patching effect are \textit{restoration probability} and \textit{logit difference}. Let $L^{*}$ and $L^{'}$ denote the logits from the corrupted and patched runs, and let $\tau, \tau^{\text{inc}}$ denote the correct/incorrect multiple-choice answer. Logit difference measures the indirect impact patching has on the model's logits. Let $L(\tau,\tau^{\text{inc}}) = \text{Logit}(\tau) \text{-} \text{Logit}(\tau^{\text{inc}})$. The logit difference is defined as $L^{'}(\tau,\tau^{\text{inc}}) \text{-} L^{*}(\tau,\tau^{\text{inc}})$. Restoration probability measures the impact of patching clean states into a corrupt run on predicting the correct answer token: $P^{\text{'}}(\tau) \text{-} P^{*}(\tau)$. 


\section{Results}
\label{sec:results}

We begin our investigation by conducting module-wise activation patching to understand the global patterns of self-attention, cross-attention, and MLP modules under both image and text corruption. We demonstrate that SIP image corruption produces results closely aligned with STR text corruption and effectively highlights the role of middle-layer MLPs, which have repeatedly proven crucial in a wide range of tasks \cite{meng2023locating}. In contrast, Gaussian Noise image corruption only identifies the final MLP layers as producing a significant patching effect. Building on this, we use SIP to perform a fine-grained analysis of cross-attention heads in BLIP and self-attention heads in LLaVA, the critical point of multimodal integration for each model. As we identify the most significant cross-attention heads in BLIP and most significant self-attention heads in LLaVA, we categorize them into three distinct functional roles, offering deeper insights into their contributions and behaviors.

\subsection{Module-wise Activation Patching}
\label{sec:module-wise}

Module-wise activation patching reveals how different model components (i.e., attention blocks and MLPs) of VLMs respond to image and text corruption, highlighting which layers and modules (MLPs, self-attention, cross-attention) are most influential for successfully completing VQA tasks. First, we compare the effects of STR and SIP on BLIP and LLaVA, as shown in Figure~\ref{fig:blip_llava} and Appendix~\ref{app:module-wise}. Across all tasks, we observe: (1) MLP patching reveals that image corruption emphasizes the importance of middle layers, while text corruption primarily affects earlier to middle layers; (2) Under text corruption, LLaVA’s self-attention demonstrates strong localization, on the correct object token in layer 0 and layer 21, similar to the cross-attention localization on the correct answer seen in BLIP at layer 3; (3) Under image corruption, LLaVA’s self-attention attends away from the correct answer token in layers 16-25; and (4) LLaVA's earlier layers identify the correct answer token while in the later layers the emphasis shift to the instruction token ``Assistant:''.  

We evaluate the effectiveness of SIP corruption by comparing it with Gaussian-noise, the image corruption technique used in previous work \cite{Palit_2023_ICCV}. In addition, we explore semantic image corruption using generative models, which shows the NOTICE framework is generalizable beyond curated datasets like SVO-Probes. Specifically, we use in-painting Stable Diffusion \citep{Rombach_2022_CVPR} to perform semantic image corruption by generating ``corrupted'' versions of clean images that only vary on one key semantic property. For instance, generating an image of a man sitting on a chair from an image of a man sitting on sand. Figure~\ref{fig:blip_svo_noise} demonstrates the promise of using generative models to implement SIP corruption since stable-diffusion module-wise patching results are highly similar to SVO-Probes. Note that generative models may occasionally introduce undesirable image artifacts (e.g., adding extra fingers to a hand) that we cannot control. We further analyze this in Appendix~\ref{app:stable_diff}. 
While using Stable Diffusion to create SIP shows promise for future CMA applications, Gaussian noise proves unreliable. In both LLaVA and BLIP, MLP modules highlight the importance of early and middle layers, while Gaussian noise only highlights the last layer (see Figure \ref{fig:blip_svo_noise} and Appendix~\ref{app:smp_gn}). Several studies highlight the importance of middle layers in transformers \cite{durrani2020analyzing, meng2023locating}, and show that MLPs store syntactic information and task-specific representations often peak in middle layer MLPs \cite{durrani2020analyzing, hewitt2019structural, goldberg2019assessing, jawahar2019does}. Additionally, self-attention patching in SIP assigns the highest probability to the correct token, unlike Gaussian noise. SIP and STR also display consistent activation patterns \textemdash long horizontal regions in MLPs (aligning with \cite{meng2023locating}) and vertical regions in self-attention \textemdash whereas Gaussian Noise shows clear differences. These findings suggest that SIP is a more effective tool for probing VLMs and identifying key components and interactions. 

\begin{figure*}[h!]
    \centering
    \includegraphics[width=\textwidth]{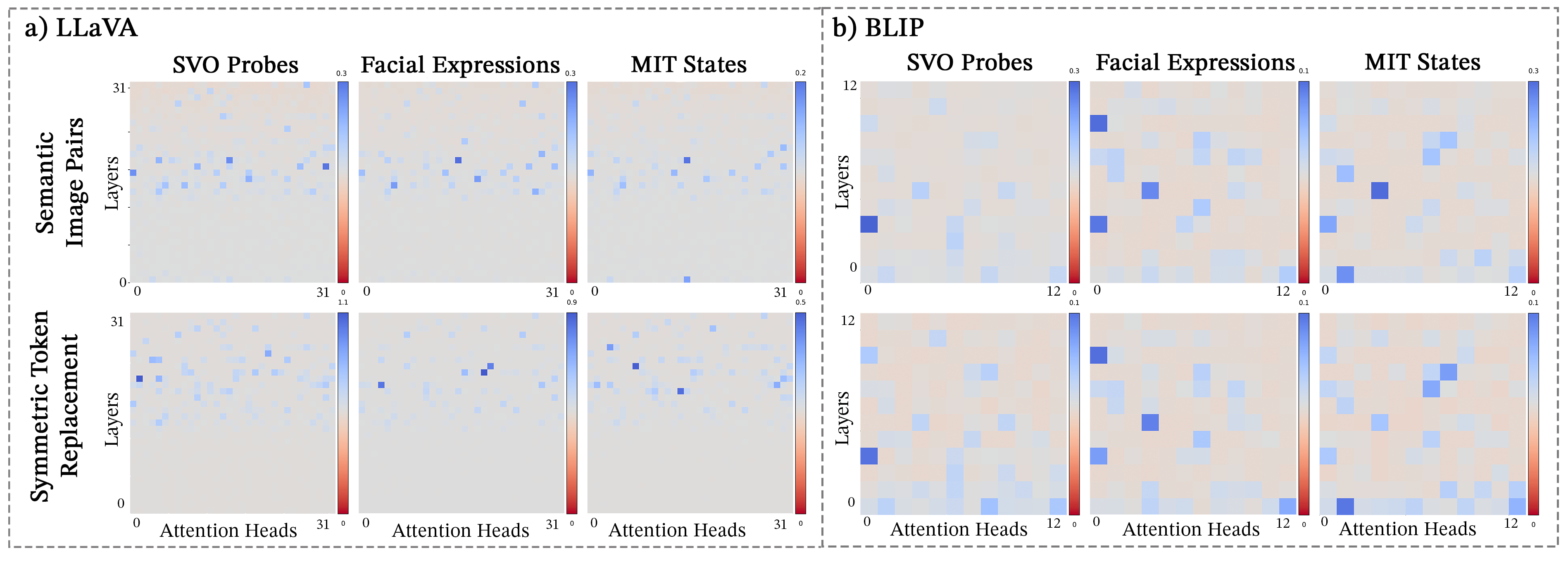}
    \caption{Logit difference demonstrating the impact of patching the correct answer for each LLaVA self-attention head and BLIP cross-attention head on the SVO-Probes, MIT States, and the Facial Expressions datasets. Many key attention heads overlap in importance across both modalities.}
    \label{fig:logit_diff}
\end{figure*}

\subsection{Universal Attention Heads Exist Across Corruption Schemes and Tasks}

\begin{table}[h!]
\centering
\begin{adjustbox}{width=\columnwidth}
\begin{tabular}{lccc}
\hline
\textbf{Model} & \textbf{Vision Heads} & \textbf{Multimodal Heads} & \textbf{Text Heads}\\ 
\hline
LLaVA & L15.H5, L16.H18 & L28.H7, L31.H27 & L22.H16, L29.H19\\
      & L14.H4, L15.H8 & L16.H24, L16.H24 & \\
      & L18.H26, L17.H0   & L19.H15, L18.H30  & \\
      & L17.H22, L14.H28  &L18.10, L21.H30 & \\
      &  L19.H4, L15.H14 & & \\
      & L17.H13, L20.H28 & & \\
\hline
BLIP  & L5.H3 & L3.H0 & L0.H11 \\
\hline
\end{tabular}
\end{adjustbox}
\caption{Universal attention heads in LLaVA and BLIP, identified as having consistently high logit differences across tasks and modality corruptions. In LLaVA, most critical self-attention heads serve multimodal or vision-specific roles. In contrast, BLIP’s dual-stream design results in distinct vision, text, and multimodal heads, supporting more modular cross-modal integration.}
\label{tab:heads}
\end{table}

Following the module-wise analysis, we focus on the attention heads, investigating their role in facilitating vision-language integration. We calculate the logit difference for each cross-attention head in BLIP and self-attention head in LLaVA by analyzing the impact of patching the correct answer across the SVO-Probes, MIT States, and Facial Expressions datasets in BLIP and the ``Assistant:'' token in LLaVA . We perform activation patching on the ``Assistant:'' token since Figure~\ref{fig:blip_llava} demonstrates that patching the ``Assistant:'' creates the largest module-wise patching effect. See Appendix~\ref{app:llava_patching_details} for more details, as well as for the patching of the correct answer token in each self-attention head in LLaVA. Figure~\ref{fig:logit_diff} demonstrates that, in BLIP, image corruption has a greater impact on logit difference compared to text corruption in SVO-Probes and MIT-States. Specifically, image corruption results in up to 20\% logit difference, while text corruption causes a difference of 6\%, suggesting BLIP relies more on visual input than text for predicting the correct answer. In LLaVA, Figure~\ref{fig:logit_diff} shows the logit difference is consistent for both modalities. These results contrast with previous findings that VLMs are more influenced by textual priors \cite{are_vl_transformers_multimodal}. Despite the difference in scale in BLIP, many critical attention heads identified under text corruption are also crucial when images are corrupted, with this overlap persisting across tasks. To quantify the overlap between the two modalities' corruptions and across tasks, we identify three universal cross-attention heads in BLIP (L0.H11, L3.H0, and L5.H3) and seven universal multimodal self-attention heads in LLaVA (L28.H2, L31.H27, L18.H10, L16.H24, L21.H30, L19.H15, and L18.H30) where L$x$.H$y$ denotes layer $x$, head $y$. 

Table \ref{tab:heads} shows ``universal attention heads'', which we defined as consistently having a logit difference two standard deviations above the average regardless of the task or the modality corruption. In LLaVA, we find only find two``text only'' heads; a majority of important self-attention heads either overlap with multimodal heads that are important for both modalities or emerge under image corruption (``vision only'' heads). This suggests visual inputs still influence the prompt, even under text corruption. We hypothesize this is due to LLaVA's early fusion, where image tokens are embedded into the textual prompt. In contrast, BLIP has exactly one vision, one text, and one multimodal head, likely because its dual-stream cross-attention design allows the model to fuse visual and textual information more effectively without requiring multiple specialized heads for different modalities. These findings highlight key differences in cross-modal integration between the two models.


\begin{figure*}[h!]
    \centering
    \includegraphics[width=\textwidth]{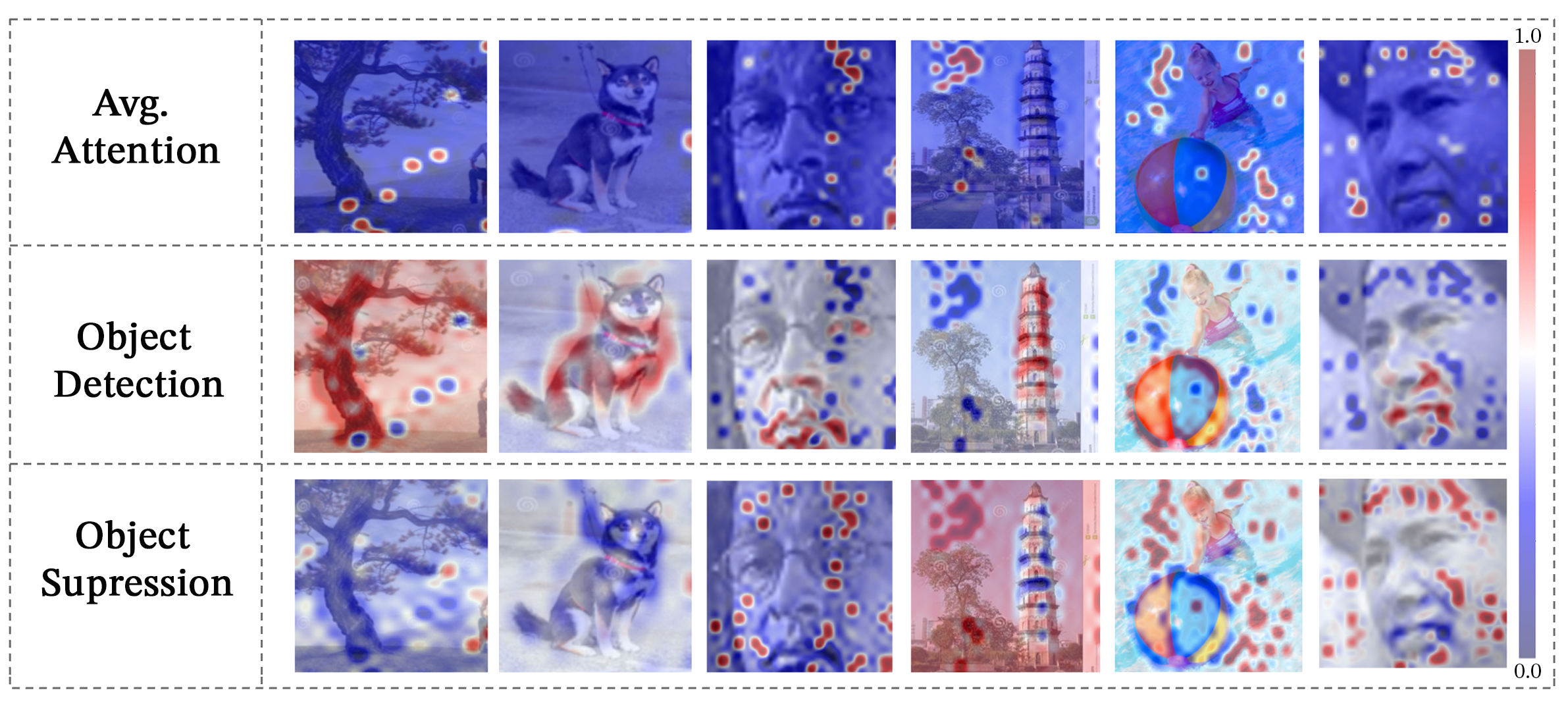}
    \caption{Universal cross-attention heads in BLIP implement object detection and object suppression. We visualize the average of all attention heads compared to the universal heads performing each function.} 
    \label{fig:catt_functions}
\end{figure*}

The presence of overlapping attention heads when patching different modalities and tasks is a novel discovery. Since cross-attention heads in BLIP and self-attention heads in LLaVA integrate vision and language, overlapping heads suggest that vision and language convey similar information to the VLM despite being distinct feature spaces. This challenges the longstanding notion that text is the dominant modality for VLMs. Additionally, the reuse of these components across tasks offers insights into model behavior and model editing, enabling targeted improvements that enhance performance, address bias, and better adapt the model to new tasks.


\subsection{Universal Attention Heads Perform Distinct Functions}
\label{sec:distinct_functions}
\begin{figure}[h!]
    \centering
    \includegraphics[width=\columnwidth]{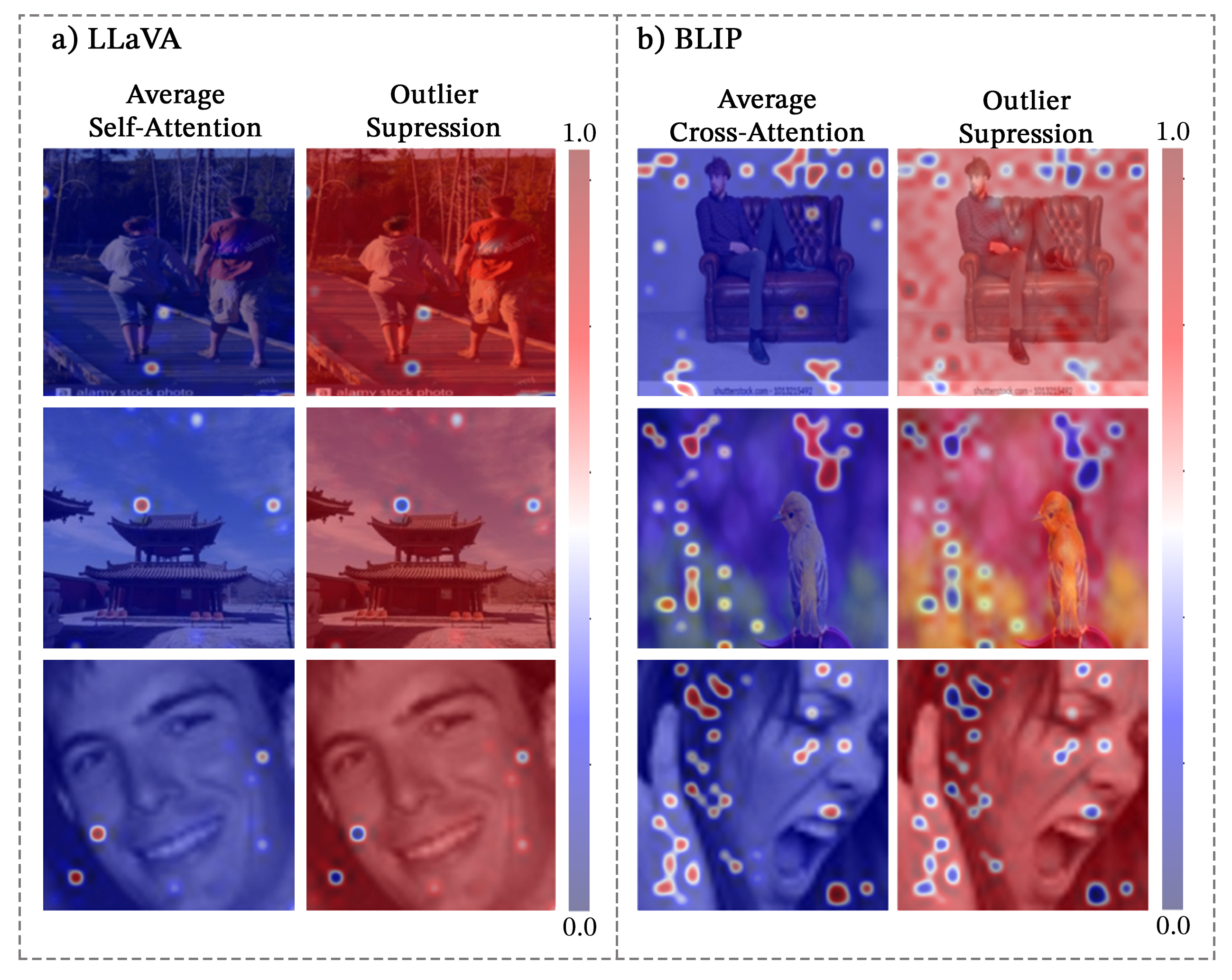}
    \caption{Universal heads in BLIP and LLaVA perform outlier suppression. We visualize the average of all attention heads compared to the universal heads performing outlier suppression.} 
    \label{fig:outlier_supp}
\end{figure}

To better understand the role of universal attention heads in multimodal integration, we visualize the cross-attention patterns between the ``correct answer'' text token and image patches in BLIP. For LLaVA, we visualize self-attention patterns of universal multimodal self-attention heads from the ``Assistant:'' token to image patches. We find that universal cross-attention heads in BLIP fall into three function classes: \textit{object detection, object suppression}, and \textit{outlier suppression}, seen in Figure~\ref{fig:catt_functions} whereas universal multimodal self-attention heads in LLaVA primarily implement outlier suppression seen in Figure ~\ref{fig:outlier_supp}.

\textbf{Outliers in Attention Patterns.} Similar to \citet{vit_registers}, we find that cross-attention patterns in BLIP and text-to-image self-attention patterns in LLaVA are dominated by ``outlier features'' with extremely high average attention scores. These outliers appear in Figure~\ref{fig:catt_functions} as non-informative image patches with dark red scores. Unlike in ViTs, where outliers are thought to obscure interpretable features in vision attention, we find attention functions reduce attention to these outliers. In BLIP, this enables object detection by linking text tokens to the relevant image patches. Our finding that outliers play an important role in VLM representations is similar to recent works showing outliers in LLMs store task-specific information \cite{outliers-task-speciic}.

\textbf{Object Detection.} In BLIP, cross-attention layer 3, head 0 (L3.H0) implements object detection by linking a text token with relevant image patches. These heads associate physical objects in images with the nouns in SVO-Probes and MIT States. Notably, L3.H0 can segment images even when the word does not explicitly appear in the image. For example, in the Facial Expressions dataset, L3.H0 implicitly associates the correct emotion (e.g., ``anger'') with the mouth region. L3.H0 is the only universal cross-attention head identified as ``multimodal,'' consistently demonstrating its importance across all tasks, regardless of modality corruption. Its ability to map text tokens to image patches highlights its critical role in fusing vision-language information and demonstrates its true multimodal capabilities.



\textbf{Object Suppression.} Cross-attention head L5.H3 in BLIP explicitly attends \textit{away} from the image patches associated with an input token and attends strongly to the outlier cross-attention patches. Object suppression cross-attention heads perform the inverse of object detectors, as they explicitly reduce attention to specific objects in an image. In SVO-Probes, this amounts to attending away from subjects or objects, and in MIT-States, L5.H3 attends away from relevant portions of the image that form the noun that is changing states. However, L5.H3 does not perform object suppression in Facial Expressions. Instead, L5.H3 performs the closely related function of outlier suppression. L5.H3 in BLIP provides a rare example of an \textit{interpretably polysemantic} attention head that implements different interpretable functions depending on the input \cite{successor_heads}.

\textbf{Outlier Suppression.} Outlier suppression heads reduce attention to high-probability image patches and distribute attention uniformly across remaining patches. All universal multimodal attention heads in LLaVA implement outlier suppression. In BLIP, L0.H11 implements outlier suppression across all tasks in this paper. For BLIP, L0.H11 is a text-only universal cross-attention head, meaning it does not produce a significant patching effect when images are corrupted. Due to its specialization, L0.H11 does not contribute to the semantic association between text tokens and image patches. Instead, it plays a crucial role in filtering out irrelevant visual information, enhancing the model's robustness.

\section{Conclusion}
In this work, we study the roles of cross-attention and self-attention in multimodal integration. To accomplish this, we introduce NOTICE, a Gaussian-Noise-free Text-Image Corruption and Evaluation pipeline for mechanistic interpretability in VLMs. Using our SIP image corruption framework and STR text corruption, NOTICE enables causal mediation analysis across both modalities. Experiments across three tasks reveal ``universal attention heads'' with consistent patching effects across BLIP and LLaVA. We show that cross-attention supports three functions: object detection, object suppression, and outlier suppression. However, self-attention primarily handles outlier suppression, indicating cross-attention’s unique role in image grounding. While attention presents a plausible, but not necessarily faithful rationale for model prediction \cite{wiegreffe2019attention}, our findings show patterns of attention heads that consistently, and causally, influence model logits. With the support NOTICE, applications using VLMs can benefit from more transparent and adaptable models, facilitating deeper insights and enhanced performance. 

\section{Limitations}

The NOTICE framework, combined with SIP corruption, offers valuable insights into the interpretability of vision-language models. In this paper, we applied our dual corruption scheme to three diverse datasets and experimented with generative models for image corruption. However, many more datasets remain unexplored. For instance, shape-based datasets like ShapeWorld \cite{kuhnle2017shapeworld} and CLEVR \cite{johnson2017clevr} allow variations in color and shape for corruption. Similarly, icon datasets like IconQA \cite{lu2021iconqa} and Emojis \cite{kralj2015sentiment}, as well as specialized datasets such as DeepFashion \cite{liu2016large}, Caltech-UCSD Birds-200-2011 \cite{wah2011caltech}, and the Stanford Cars dataset \cite{kramberger2020lsun}, can be turned into ``clean'' and ``corrupt'' pairs. In future work, we will expand our experiments to include a wider range of datasets and models, further enhancing the robustness and generalizability of our findings.

\bibliography{anthology,custom}

\appendix

\newpage

\section{VQA-Prompts}
\label{app:vqa_prompts}

For each task, we create a VQA prompt that asks the model to choose between a correct answer, which corresponds to the image, and an incorrect answer that corresponds to a semantically minimal pair. While in SVO-Probes, each subject, verb, or object, came with it's own semantically different counterpart, in MIT-States dataset and Facial Expression dataset, we constructed such counterparts manually. In MIT-States, we separated states into color, shape, and texture, and selected counterparts for each question from the respective categories. We grouped similar states (e.g., ``huge'', ``large'') together, and ensure that the selected counterpart would not be similar in meaning to the correct answer. Similarly, in the Facial Expressions dataset, we grouped the seven emotions into positive, negative, and neutral, and selected counterparts from opposing emotions, to ensure the model can correctly chose one answer. Recall that the SVO-Probes dataset consists of subject, verb, and object triplets. Each triplet is presented in singular form and non-conjugated. We use GPT-3.5 Turbo to perform grammatical error correction on all input prompts in SVO-Probes while minimally perturbing the template structure. 
We evaluate BLIP using exact match accuracy, which requires the generated answer to match the correct answer exactly, including capitalization and conjugation. BLIP achieves 66.1\% accuracy on the facial expressions dataset, 44.6\% on MIT States, and 16.1\% on SVO-Probes (see Table \ref{tab:accuracy_by_dataset} for category-wise accuracy). In our experiments, we use samples that BLIP accurately predicted. For each task, we select 500 samples across all categories: subjects, verbs, and objects for SVO-probes; colors, shapes, and textures for MIT-States; and one of the five emotions BLIP correctly identified for the facial expression identification task. To mitigate model bias toward the position of the correct answer in multiple-choice prompts, we equally distribute 250 samples where the correct option precedes and 250 where it follows the ``or'' token.

\section{Datasets}
\label{app:datasets}
\textbf{SVO-Probes.} The Subject-Verb-Object (SVO) Probes dataset \citep{hendricks2021probing} consists of over 48,000 image-language pairs that vary on exactly one of the subject, verb, and object tuples that describe an image. For example, Figure~\ref{fig:corruption_scheme} depicts image-text pairs with the captions ``A child crossing the street'' (child, crossing, street) and ``A lady crossing the street'' (lady, crossing, street). In this instance, the two image pairs only vary on the subject of the image, but the verb (crossing) and the object (street) are identical. SVO-Probes tests if multimodal models can capture fine-grained linguistic information about the content of an image. 

\textbf{MIT-States.} The MIT States dataset \cite{isola2015discovering} is a collection of 245 object classes, 115 attribute classes, and over 53,000 images. The dataset is used to study how models can generalize state transformations across different object classes. For instance, learning what ``melted'' looks like on butter can help recognize ``melted'' in chocolate. 

\textbf{Facial Expressions.} The Facial Expression Recognition 2013 (FER-2013) \cite{goodfellow2013challenges} dataset includes 35,887 grayscale images of faces resized to 48x48 pixels. The images are categorized into seven emotion classes: anger, disgust, fear, happiness, sadness, surprise, and neutral. The FER-2013 dataset is valuable for comparing feature learning methods with hand-engineered features in emotion recognition tasks.

\section{BLIP and LLaVA Performance}

\begin{table}[h!]
\centering
\resizebox{\columnwidth}{!}{%
\begin{tabular}{lcc}
\hline
\textbf{Dataset} & \textbf{BLIP Accuracy (\%)} & \textbf{LLaVA Accuracy (\%)} \\ 
\hline
SVO & 16.10 & 40.89 \\ 
Probes    &  &  \\ 
\hline
MIT & 44.60 & 67.81 \\ 
States      &  &  \\ 
\hline
Facial  & 66.10 & 79.31 \\ 
Expressions        &  &  \\ 
\hline
\end{tabular}
} 
\caption{Comparison of BLIP and LLaVA overall accuracy by dataset, expressed as percentages.}
\label{tab:accuracy_by_dataset}
\end{table}

In evaluating the BLIP and LLaVA, we focused on exact match accuracy, which necessitates that the model's generated answer must exactly match the correct answer in terms of both content and form, including specifics like capitalization and conjugation. This stringent criterion ensures that the model's understanding and response generation align closely with the intended semantics of the test prompts. Table \ref{tab:accuracy_by_dataset} shows the performance of BLIP and LLaVA (zero-shot) on each task, and each category within the task. 

\section{SIP with Stable Diffusion}\label{app:stable_diff}

With the primary goal of establishing a more controlled environment conducive to the generation of semantically similar corrupt (negative) images from clean(positive) images, we also explore the use of stable diffusion to produce images that are similar to clean images from the SVO-Probes dataset, with the sole variation being in only one among the subject, verb, or object of the image. 

Our experiment design utilizes the \textit{stable-diffusion-2-inpainting} \citep{Rombach_2022_CVPR} to generate a corrupt image using the clean image as a base template. In this context, we define a prompt structure for the model as “a picture of $\text{sub}_{neg} + \text{verb}_{neg} + \text{obj}_{neg}$” where $<\text{sub}_{neg}, \text{verb}_{neg}, \text{obj}_{neg}>$ denote the svo-triplet for the negative image.  

For the parametric constraints on the diffusion model pipeline, we utilize a DDIM Scheduler \citep{song2022denoisingdiffusionimplicitmodels}, setting the DDIM value $=120$ and the prompt guidance scale value to $8$ for all the image generations. For generating images on the template of the base positive image, we use a contour mask, which allows for inpainting of the negative subject, verb, or object into the positive image. In such cases, a variation in quality is observable in the images introduced due to the probabilistic nature of the model. 

\begin{figure*}[h!]
    \centering
    \includegraphics[width=0.5\textwidth]{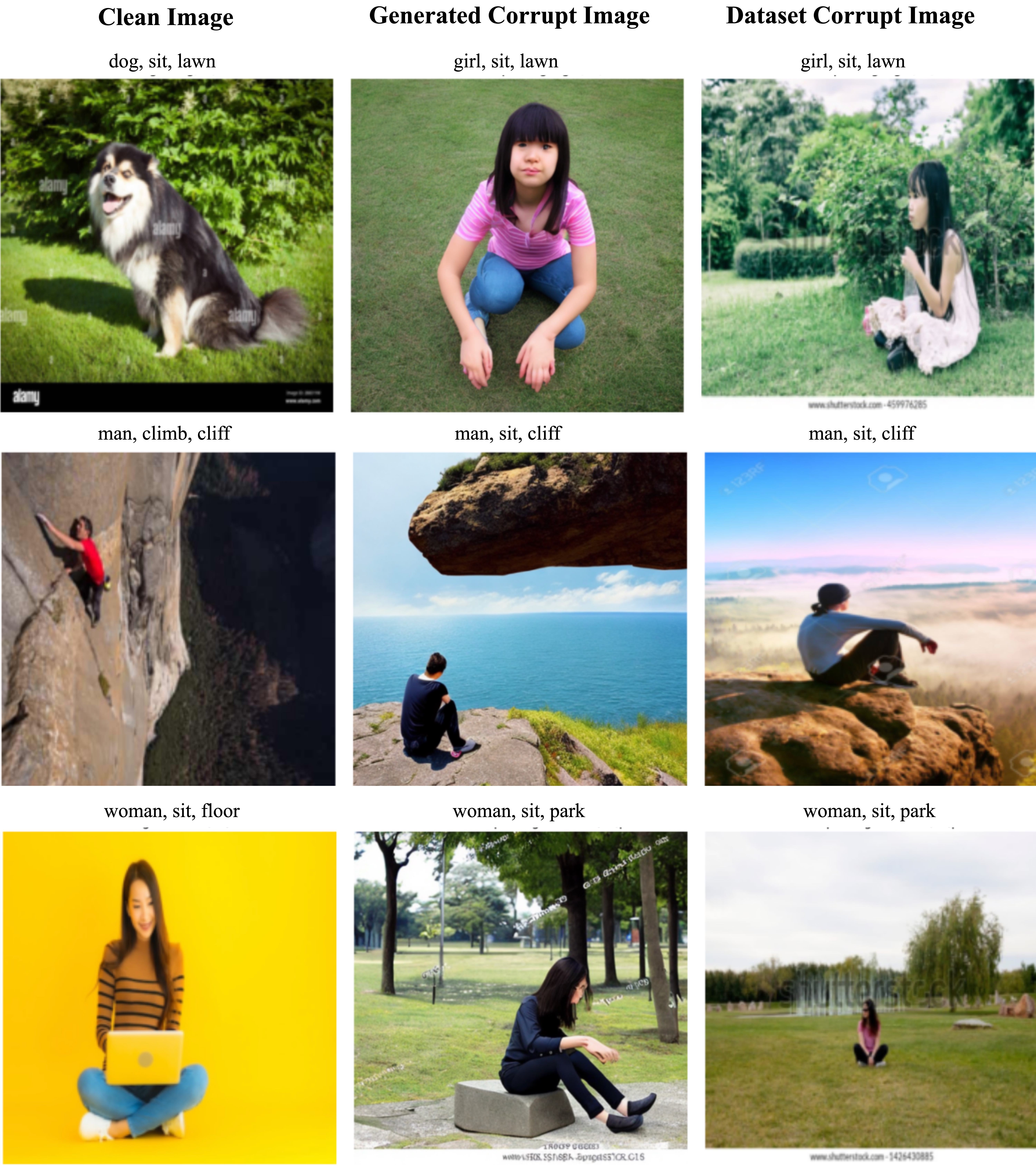}
    \caption{Successful examples of images generating using \textit{stable-diffusion-2-inpainting} that showcase corrupt images close enough to the clean image yet different in (1) Subject (2) Verb (3) Object}
    \label{fig:sd-images-pos}
\end{figure*}

\begin{figure*}[h!]
    \centering
    \includegraphics[width=0.5\textwidth]{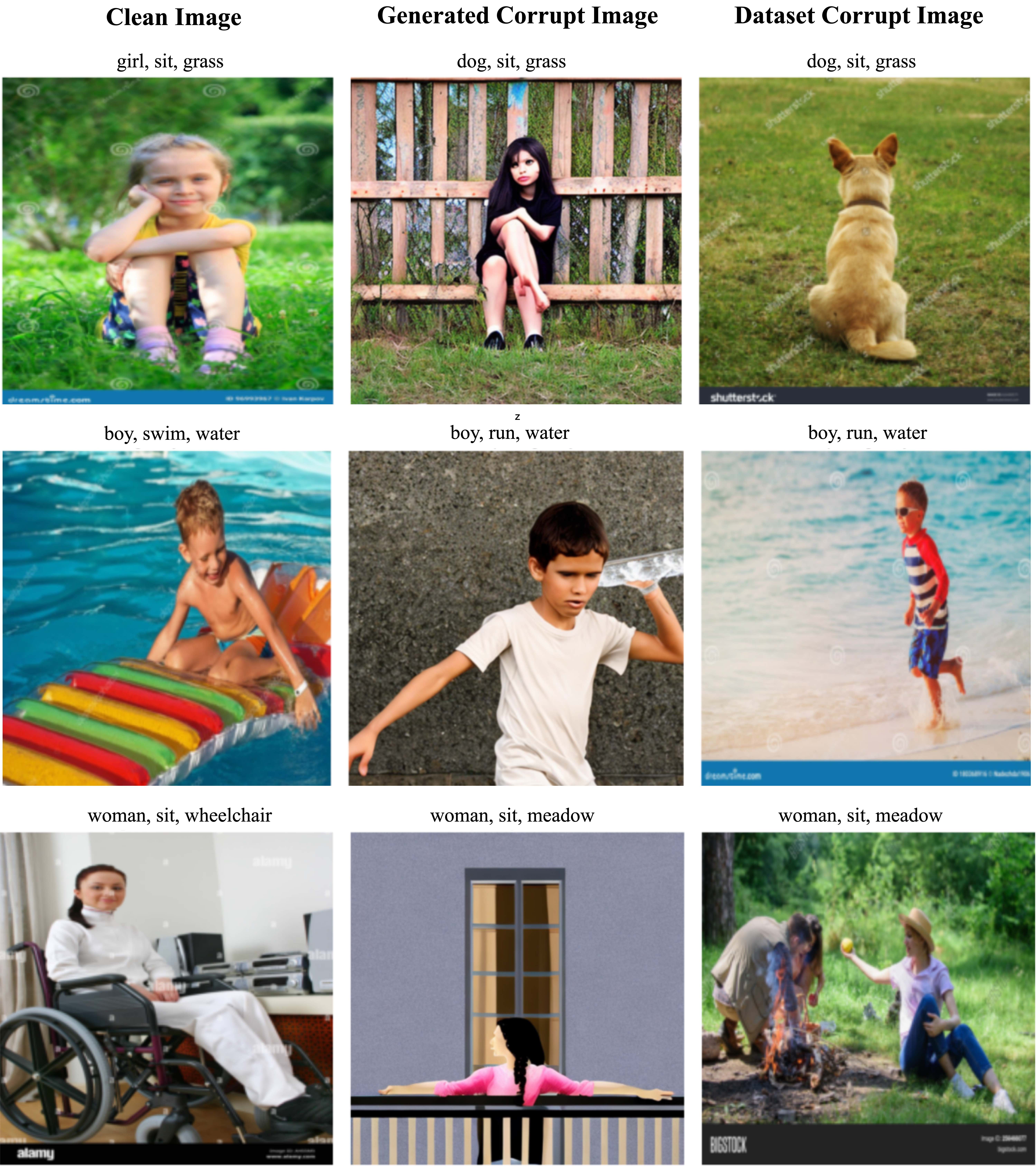}
    \caption{Poor examples of images generating using \textit{stable-diffusion-2-inpainting} that showcase corrupt images which deviate from the clean image  in (1) Subject (2) Verb (3) Object}
    \label{fig:sd-images-neg}
\end{figure*}

Holistically, the generated images from \textit{stable-diffusion-2-inpainting} are largely closer to the original corrupt images from SVO-Probes dataset where the image annotations utilize simpler terminologies to depict any one of the three $\text{sub}_{neg},\text{verb}_{neg},\text{obj}_{neg}$ such as ``grass'' instead of ``meadow'' (refer to Fig. \ref{fig:sd-images-pos} and \ref{fig:sd-images-neg}). 

\section{Compute Resources}
For each experiment, we use one NVIDIA GeForce RTX 3090 GPU. Running BLIP for each activation patching experiment ranged from 12 hours for SVO probes to 10 hours in MIT states and Facial Expressions. Running LLaVA for each activation patching experiment ranged from 96 hours for SVO probes to 90 hours in MIT states and Facial Expressions.

\section{Results}
\subsection{SIP vs. Gaussian Noise Image Corruption.}\label{app:smp_gn}

\begin{figure*}[h!]
    \centering
    \includegraphics[width=\textwidth]{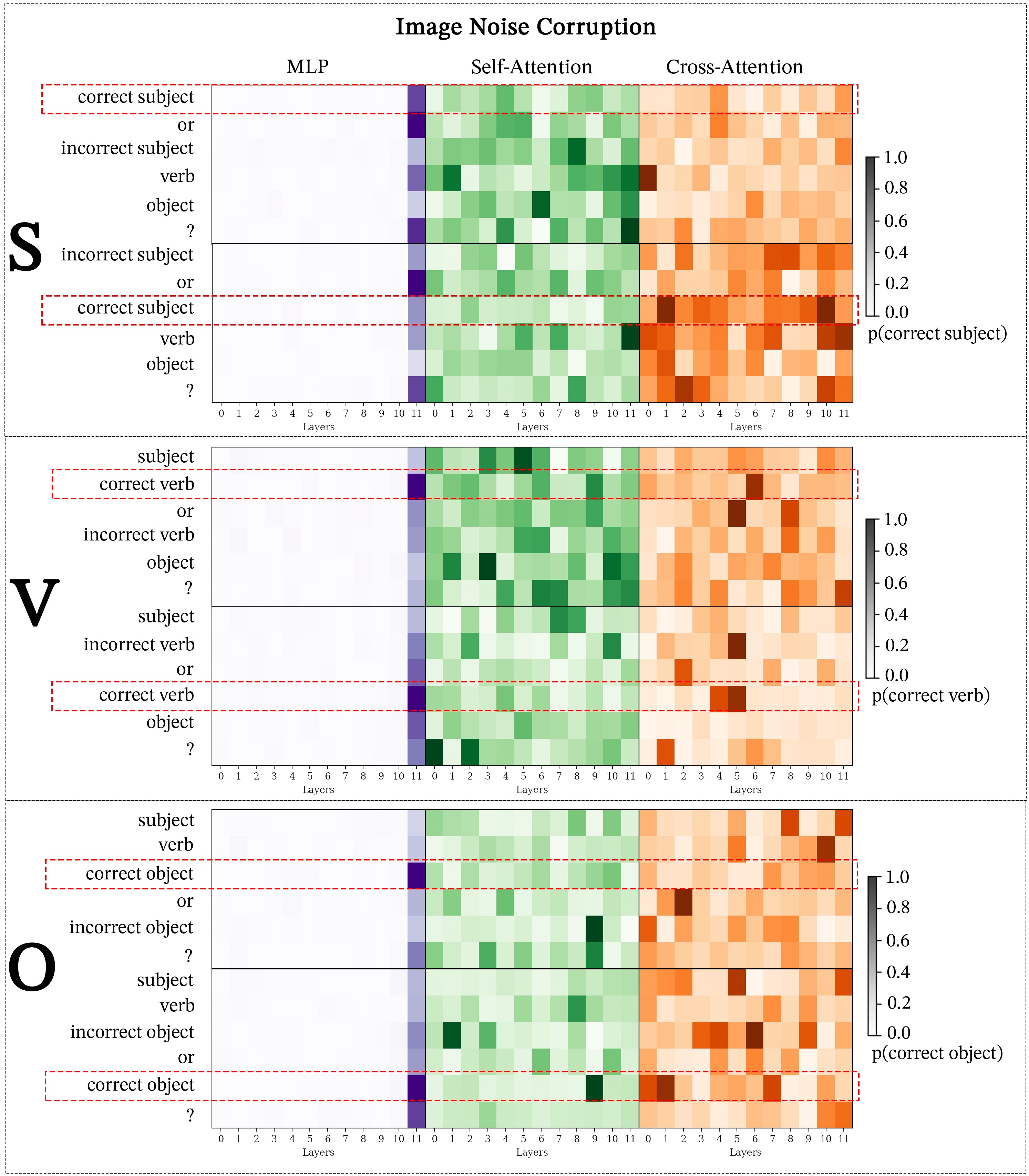}
    \caption{Comprehensive view of Gaussian noise effects across all tasks in the SVO-Probes dataset, providing a contrast to SIP effects and highlighting the differential impact of noise on model accuracy and layer activation.}
    \label{fig:all_blip_noise}
\end{figure*}

Figure \ref{fig:all_blip_noise} illustrates the comparative effects of Semantic Image Pairs (SIP) and Gaussian Noise (GN) image corruption across all tasks, highlighting SIP’s superior ability to emphasize the significance of middle layers in the MLP, in contrast to GN's focus on later layers. These observations affirm SIP's enhanced reliability and effectiveness in revealing the nuanced functioning of vision-language models, supporting its use for in-depth interpretability studies.

\subsection{Module-wise Activation Patching}
\label{app:module-wise}
Module-wise activation patching offers a detailed examination of how different components of the BLIP model respond to image and text corruption, using Semantic Image Pairs (SIP) and Gaussian noise for images and Symmetric Token Replacement (STR) for texts. This approach is crucial for uncovering the roles of self-attention, cross-attention, and MLP modules within the architecture, particularly focusing on their reaction to varied corruption methods. Through these experiments, we aim to pinpoint the layers most critical for the model's decision-making processes and explore how different corruption strategies affect the transparency and reliability of the model's internal workings. This analysis not only tests the robustness of the SIP technique against the traditional Gaussian noise approach but also highlights the strategic layers that are pivotal in handling multimodal data integration.

\begin{figure*}[h!]
    \centering
    \includegraphics[width=\textwidth]{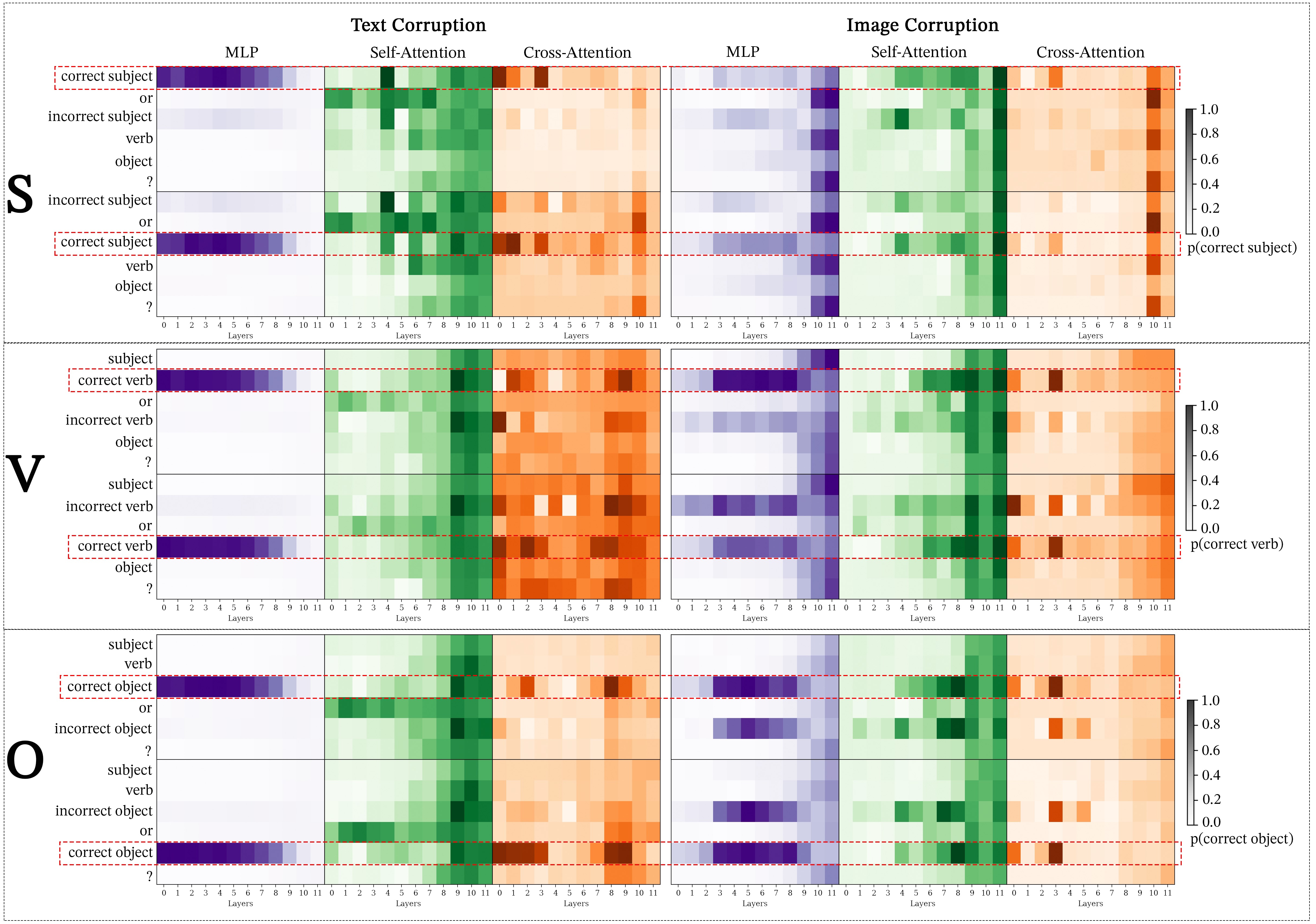}
    \caption{Module-wise activation patching results on BLIP for the SVO-Probes dataset. }
    \label{fig:blip_svo}
\end{figure*}

\begin{figure*}[h!]
    \centering
    \includegraphics[width=\textwidth]{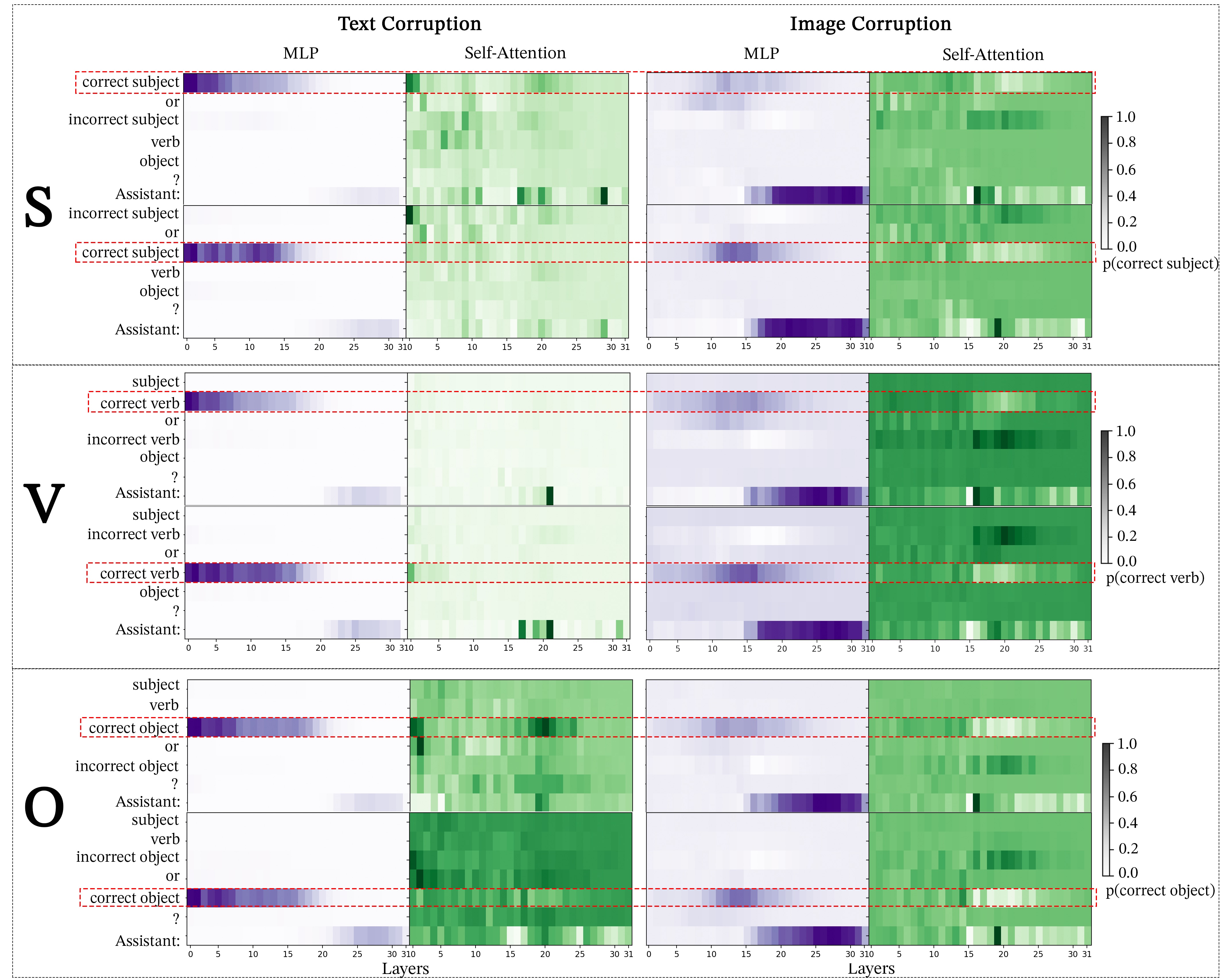}
    \caption{Module-wise activation patching results on LLaVA for the SVO-Probes dataset.}
    \label{fig:llava_svo}
\end{figure*}

First, Figure \ref{fig:blip_svo} and Figure \ref{fig:llava_svo} show module-wise activation patching for SVO Probes, broken down into subjects, verbs, and objects, VQA prompts. 

\begin{figure*}[h!]
    \centering
    \includegraphics[width=\textwidth]{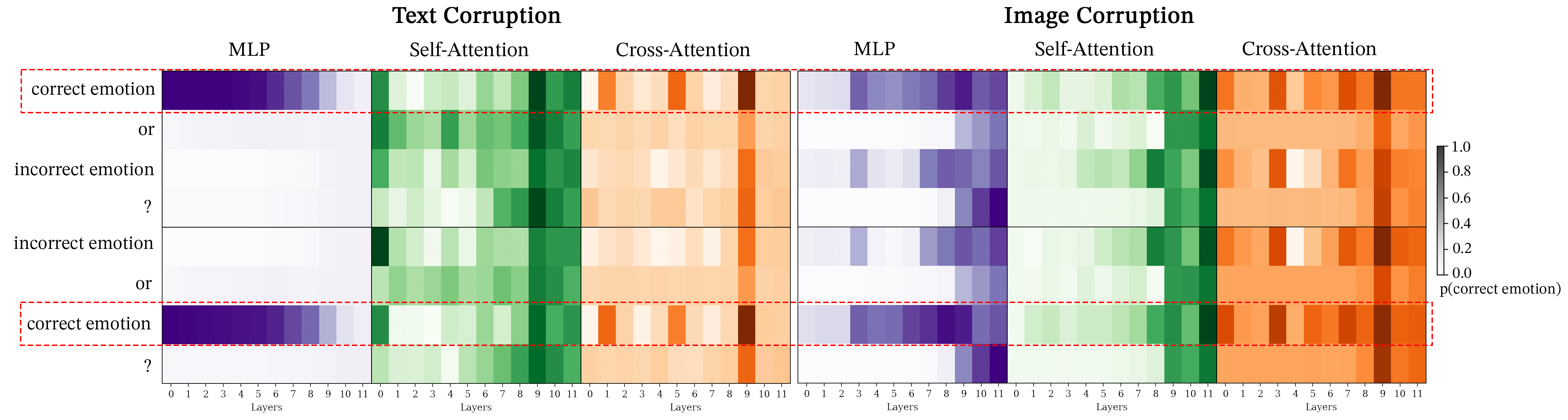}
    \caption{Module-wise activation patching results on BLIP for the Facial Expressions dataset.}
    \label{fig:blip_emotions}
\end{figure*}

\begin{figure*}[h!]
    \centering
    \includegraphics[width=\textwidth]{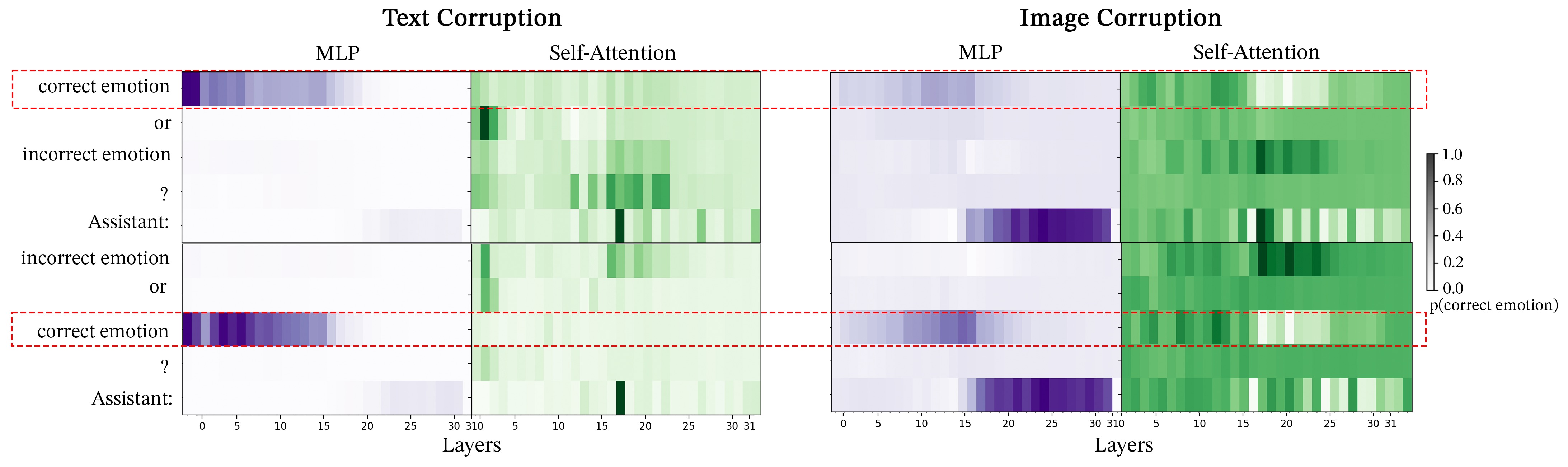}
    \caption{Module-wise activation patching results on LLaVA for the Facial Expressions dataset.}
    \label{fig:llava_emotions}
\end{figure*}

Next, we examine module-wise activation patching for the Facial Expressions dataset in Figure \ref{fig:blip_emotions} for BLIP and Figure \ref{fig:llava_emotions}. We note that image and text corruption results in similar patterns. This provides further support to SIP corruption aligning with STR corruption. 

\begin{figure*}[h!]
    \centering
    \includegraphics[width=\textwidth]{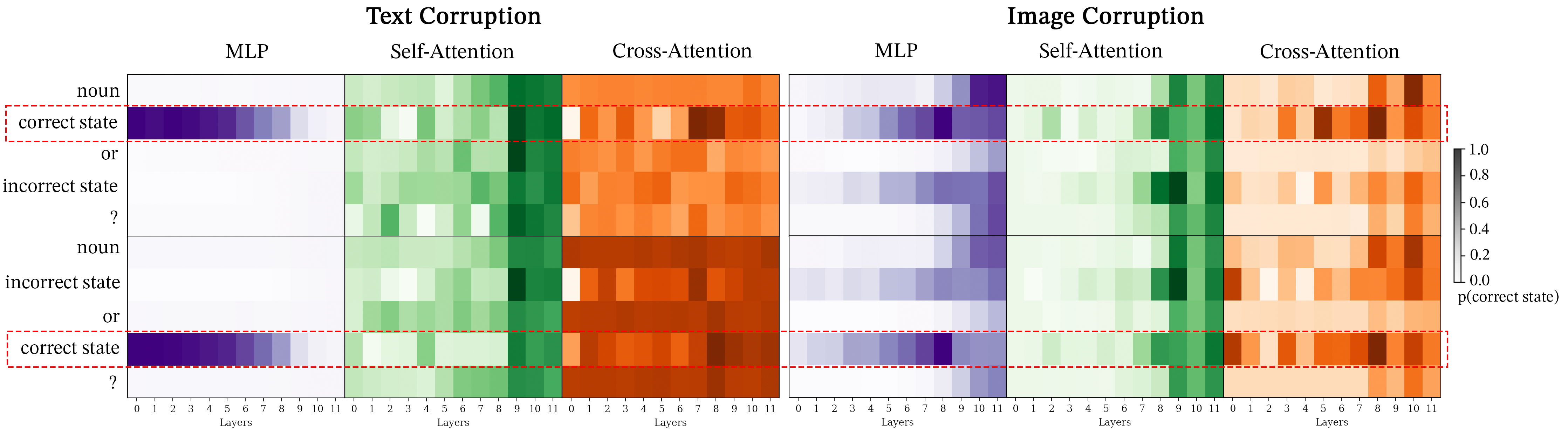}
    \caption{Module-wise activation patching results on BLIP for the MIT States dataset. }
    \label{fig:blip_mit_states}
\end{figure*}

\begin{figure*}[h!]
    \centering
    \includegraphics[width=\textwidth]{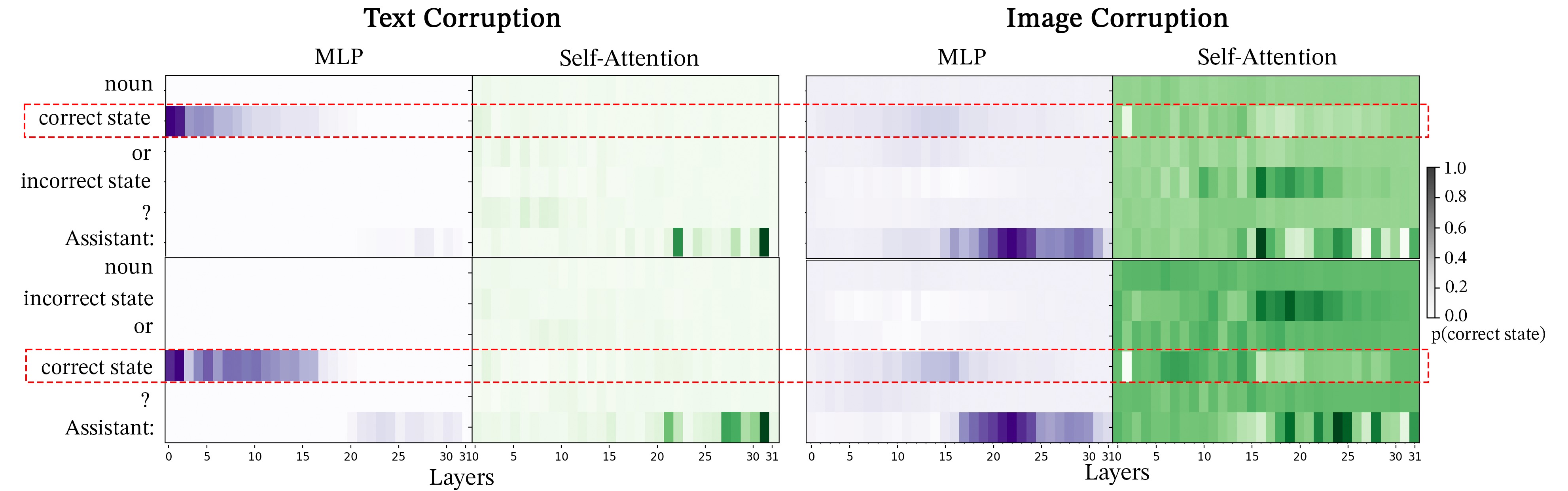}
    \caption{Module-wise activation patching results on LLaVA for the MIT States dataset. }
    \label{fig:llava_mit_states}
\end{figure*}

Lastly, Figure \ref{fig:blip_mit_states} and Figure \ref{fig:llava_mit_states} exhibit outcomes similar to the other tasks but with a notable distinction: there's an increased likelihood of accurately predicting the correct token within the cross-attention heatmaps during text corruption. This suggests that the multiple-choice VQA prompts used in the MIT States dataset may be more definitive compared to the other datasets. While distinguishing between emotions such as ``angry'' and ``sad,'' or identifying subtle differences like ``woman'' versus ``girl'' can be challenging in the Facial Expressions and SVO Probes datasets, respectively, the task of differentiating between clearly contrasting attributes like ``modern'' versus ``ancient'' buildings in MIT States appears to be more straightforward.

\begin{table}[ht]
\centering
\resizebox{\columnwidth}{!}{%
    \begin{tabular}{c|c|c|c}
        \hline 
        \hline
         Cross-Attention Functions & SVO & MIT & Facial Expressions  \\
         \hline 
        object detection & L3.H0 & L3.H0 & L3.H0 \\
        object suppression & L5.H3 & L5.H3 & N/A \\
        outlier suppression & L0.H11 &  L0.H11 & L0.H11, L5.H3 \\
        \hline 
        \hline
    \end{tabular}
} 
    \caption{Universal heads identified through cross-attention head patching, alongside their functions.} 
    \label{tab:universal_catt_functions}
\end{table}




\section{LLaVA Self-Attention Patching Details}

\begin{figure*}
    \centering
    \includegraphics[width=\textwidth]{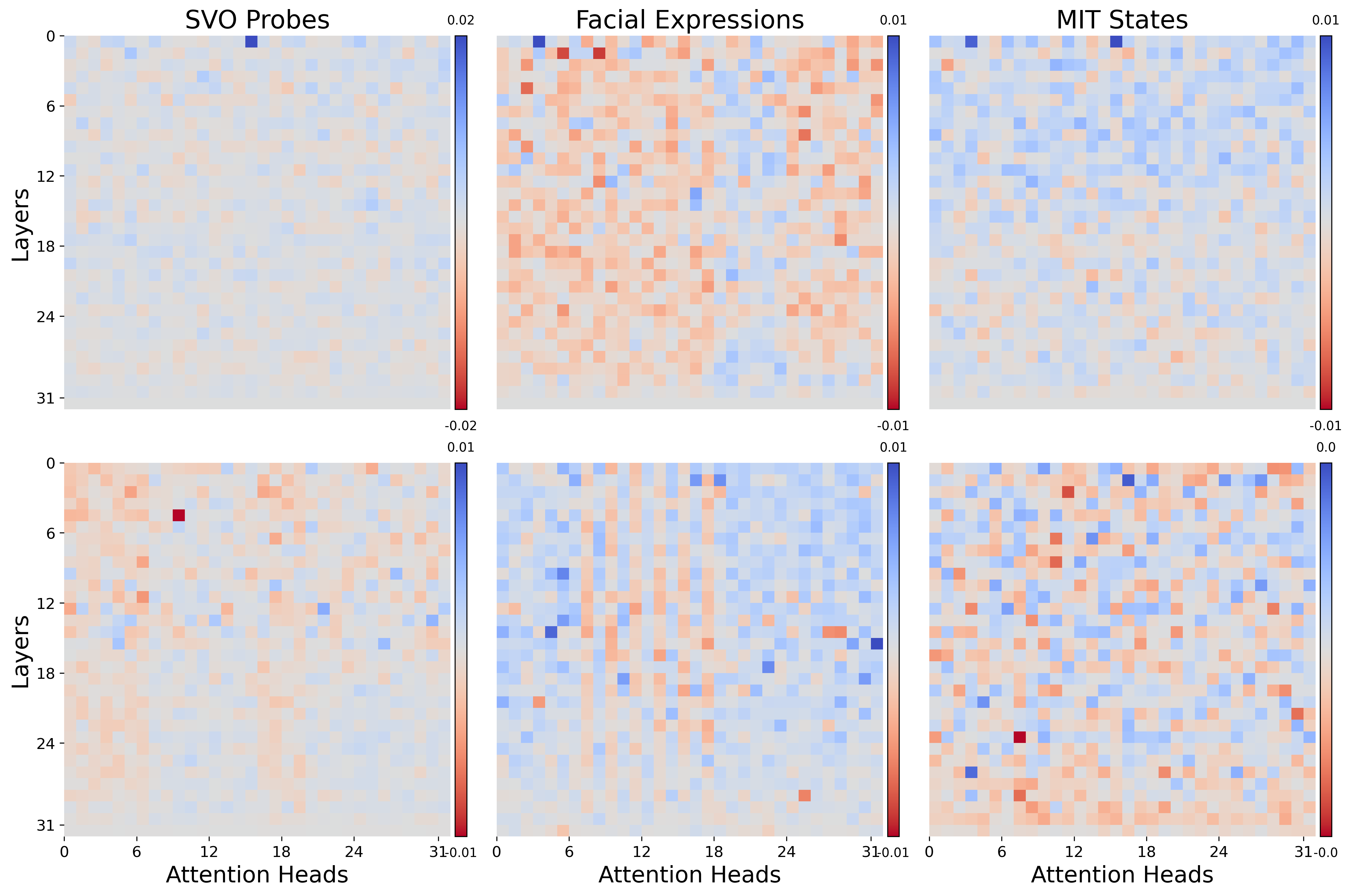}
    \caption{Logit difference when patching self-attention heads in LLaVA on the correct multiple choice option in SVO-Probes, MIT States and Facial Expressions. Top row shows SIP image corruption, the bottom row shows STR text corruption.}
    \label{fig:llava_last_token}
\end{figure*}

\label{app:llava_patching_details}
We consider two methods for performing activation on LLaVA's self-attention heads. First, we patch only on the correct answer token. Although Figure~\ref{fig:logit_diff} demonstrates a strong patching effect in BLIP, patching the correct answer token in LlaVA results in an extremely small logit difference, with the largest average logit difference being 0.016 as shown in Figure~\ref{app:llava_patching_details}, regardless of modality corruption or dataset. Patching the correct answer token does not result in any ``universal attention heads'' indicating that self-attention does not play a pivotal role in associating the relevant images patches with the correct answer token in LLaVA. Module-wise patching using image corruption for both MLPs and Self-Attention blocks shows that patching the 'Assistant:' token consistently produces the strongest patching effect for LLaVA. Accordingly, we experiment with patching the 'Assistant:' token for LLaVa's self-attention heads. Patching the ``Assistant:'' token creates a far stronger logit difference with some heads producing an average logit difference as high as 0.4. Patching 'Assistant:' produces consistent results with 6 universal self-attention heads compared to 0 when patching the correct answer token. Both the consistency and strength of patching the 'Assistant:' token highlight the importance of the 'Assistant:' token in storing information relevant to the image. Future studies performing activation patching on instruction-tuned VLMs that utilize causal self-attention to fuse modalities should focus on the information in the specific instruction tokens.   

\subsection{Comparing Cross-Attention and Self-Attention Head Functions}

Our result that the cross-attention mechanism VLMs consistently associates the nouns with their corresponding image patches and that causal self-attention does not associate nouns to image regions is supported by previous works investigating cross-attention and self-attention patterns in toy VLMs \cite{attention_as_grounding}. Although self-attention heads in LLaVA associate neither the correct answer token nor the 'Assistant:' token with the relevant image patches, universal self-attention heads in LLaVA have the primary function of mitigating outlier attention patterns in images.

Important cross-attention heads in BLIP are scattered throughout the network ranging from layer 0 to layer 9. In contrast, import self-attention heads in LLaVA are concentrated in the middle layers of the network with x / y self-attention heads between 14-18. 
\subsection{Cross-Attention Head Function Classes}
\label{app:catt_functions}

Understanding the specific functions of cross-attention heads within vision-language models (VLMs) is crucial for mechanistic interpretability. In this section, we delve into the distinct roles played by universal cross-attention heads identified by our NOTICE scheme. These heads can be categorized into three primary function classes: \textit{implicit object detection, object suppression,} and \textit{outlier suppression}. Below, we provide detailed examples of these functions along with corresponding visualizations from the SVO-Probes, MIT-States, and Facial Expressions datasets.

\textbf{L0.H11 - outlier suppression.} Universal cross-attention head L0.H11 implements outlier suppression in all tasks. It actively attends away from high-probability image patches and uniformly attends to the remaining patches. This function significantly reduces the mean of the average cross-attention outputs, thereby enhancing the model's robustness by filtering out irrelevant visual information. L0.H11 does not produce a significant patching effect when images are corrupted, highlighting its role as a text-only universal cross-attention head.

\begin{figure*}[h!]
\centering
\includegraphics[width=\textwidth]{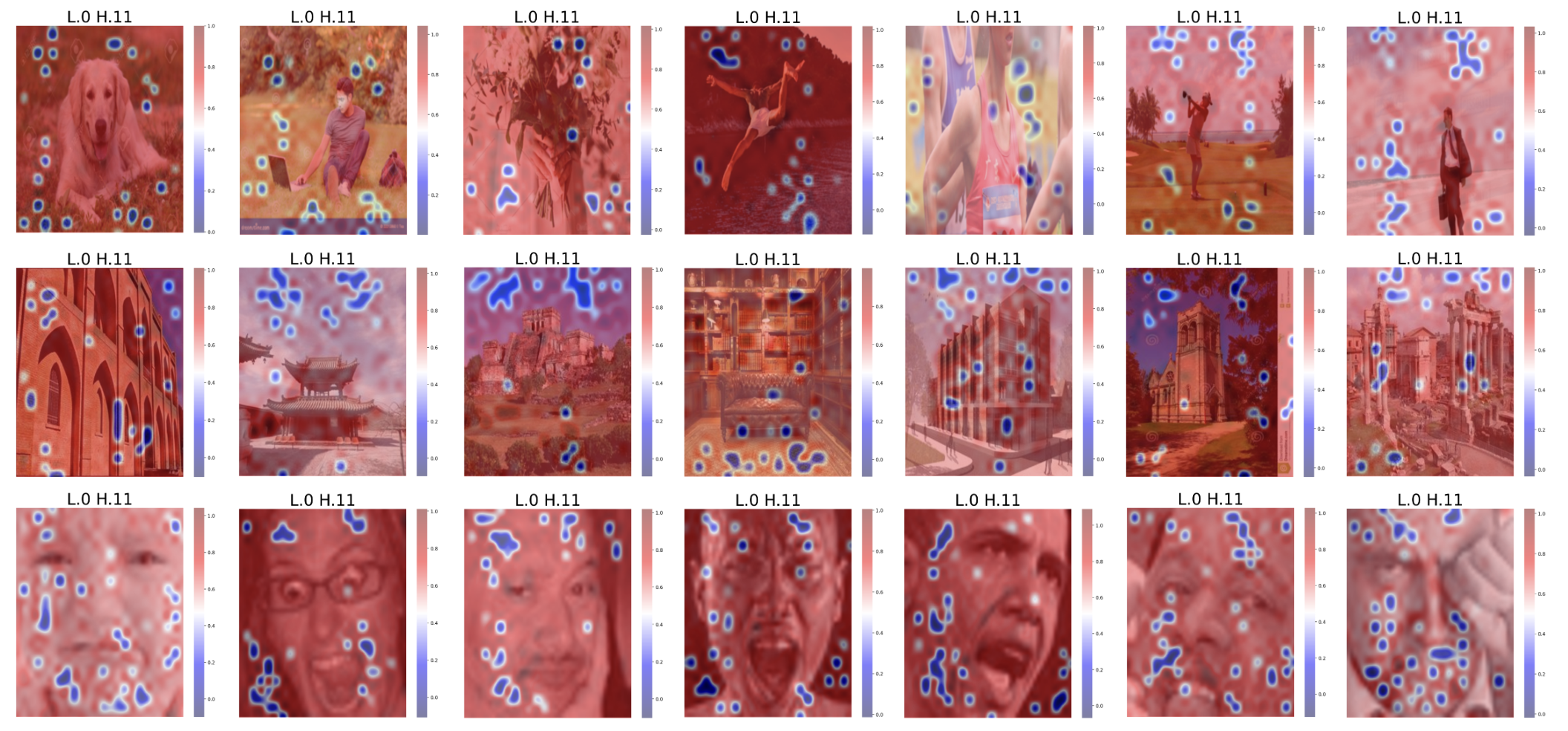}
\caption{Cross-Attention patterns for universal cross-attention head L0.H11 on a sample of images from each dataset.}
\label{fig:l0h11}
\end{figure*}

\textbf{L3.H0 - Object detection.} Cross-attention layer 3, head 0 (L3.H0) performs implicit object detection by associating text tokens with relevant image patches. It consistently demonstrates its importance across all tasks, reliably associating the correct emotion text token with the mouth region in the Facial Expressions dataset, and associating physical objects in images with their corresponding nouns in SVO-Probes and MIT-States. This head's ability to map input text tokens to associated image patches underscores its critical role in fusing vision-language information.

\begin{figure*}[h!]
\centering
\includegraphics[width=\textwidth]{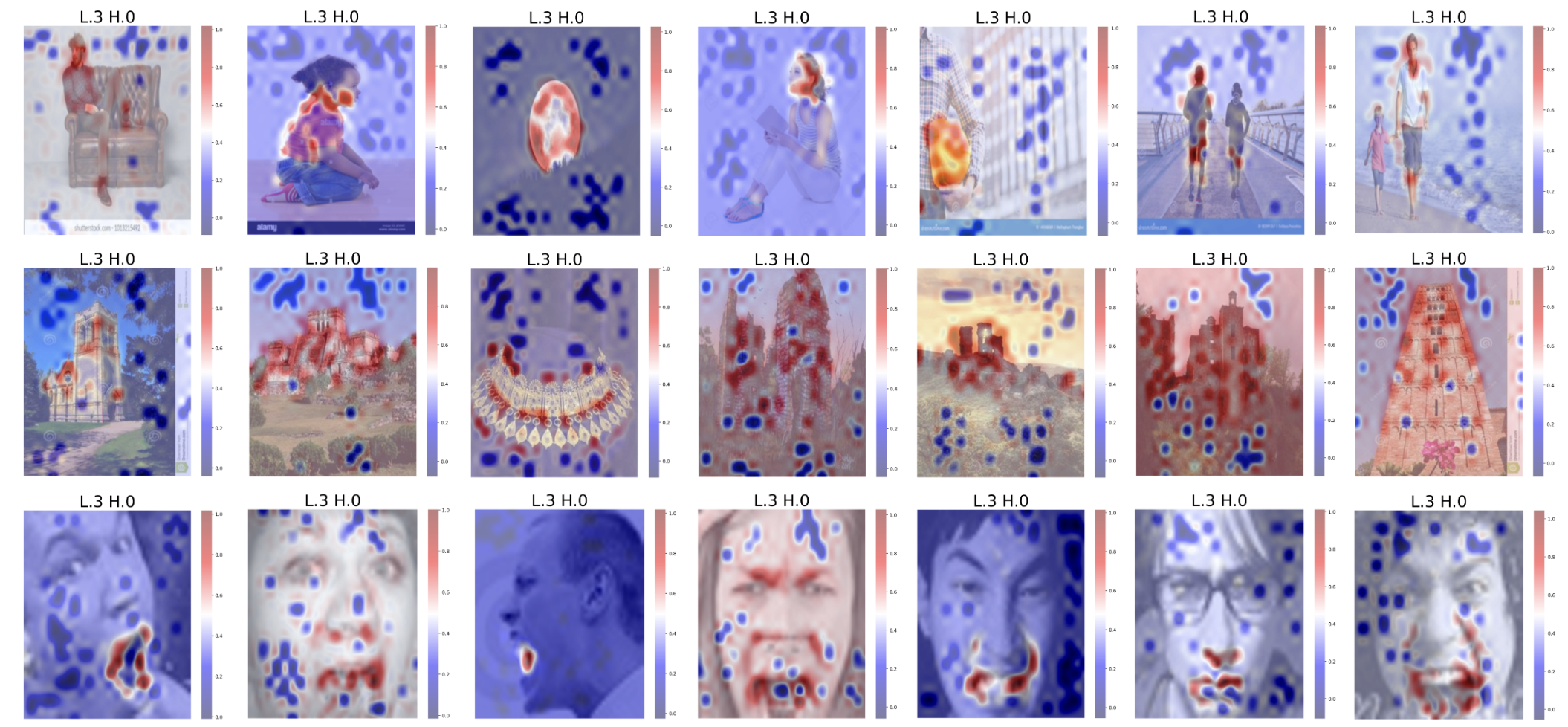}
\caption{Cross-Attention patterns for universal cross-attention head L3.H0 on a sample of images from each dataset.}
\label{fig:l3h0}
\end{figure*}

\textbf{L5.H3 - object suppression and outlier suppression.} Cross-attention head L5.H3 performs object suppression by attending away from objects associated with an input token and attending strongly to outlier patches. In SVO-Probes, it attends away from subjects or objects, while in MIT-States, it attends away from relevant portions forming the noun that is changing states. In the Facial Expressions dataset, L5.H3 performs outlier suppression instead of object suppression.

\begin{figure*}[h!]
\centering
\includegraphics[width=\textwidth]{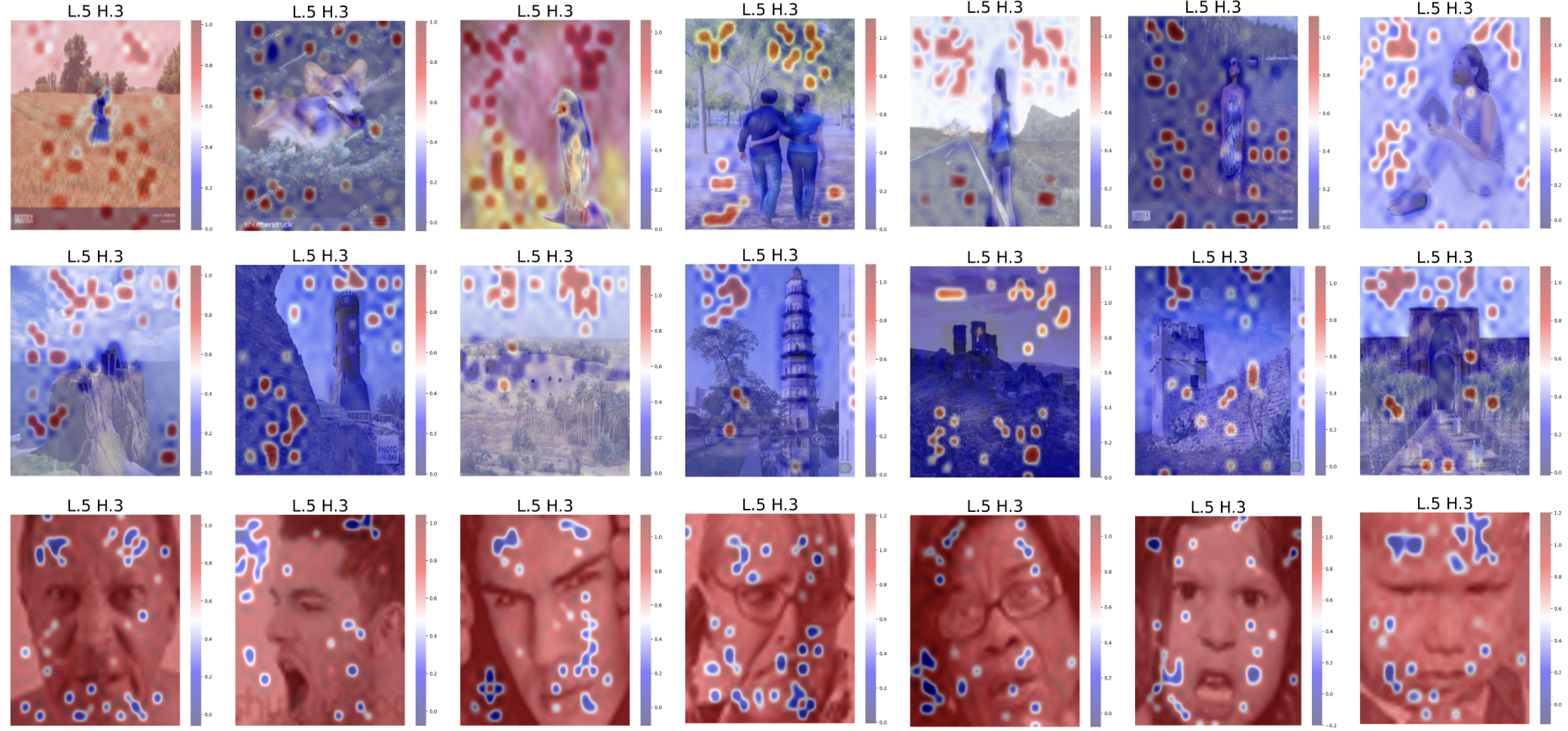}
\caption{Cross-Attention patterns for universal cross-attention head L5.H3 on a sample of images from each dataset.}
\label{fig:l5h3}
\end{figure*}

These examples underscore the diverse functions of cross-attention heads within BLIP, each playing a pivotal role in processing and integrating multimodal information. The identification and categorization of these heads provide a deeper understanding of the internal mechanisms driving VLMs, highlighting the significance of targeted model improvements and adaptations.

\end{document}